\documentclass{article}

\PassOptionsToPackage{numbers}{natbib}
\usepackage[preprint]{neurips_2019}

\usepackage[utf8]{inputenc} % allow utf-8 input
\usepackage[T1]{fontenc}    % use 8-bit T1 fonts
\usepackage{hyperref}       % hyperlinks
\usepackage{url}            % simple URL typesetting
\usepackage{booktabs}       % professional-quality tables
\usepackage{microtype}      % microtypography
\usepackage{comment} % enables the use of multi-line comments (\ifx \fi) 
\usepackage{amsmath, amssymb, amsfonts,amsthm}
\usepackage{mathtools}
\usepackage{xfrac}
\usepackage{nicefrac}
\usepackage{dsfont}
\usepackage{mathabx}
\usepackage{cleveref}
\usepackage[binary-units]{siunitx}
\usepackage{cleveref}
\usepackage{subcaption}
\usepackage{caption}
\usepackage{xspace}
\usepackage{verbatim}
\usepackage{physics}
\usepackage{algorithm}
\usepackage[noend]{algpseudocode}
\usepackage{color}
\usepackage{blindtext}
\usepackage{tabularx}
\newcommand{\calA}[0]{\ensuremath{\mathcal{A}}\xspace}
\newcommand{\calS}[0]{\ensuremath{\mathcal{S}}\xspace}

\newcommand{\discSet}[0]{\ensuremath{\mathcal{D}}\xspace}
\newcommand{\absDiscSet}[0]{\ensuremath{\abs{\discSet}}\xspace}
\newcommand{\discVal}[0]{d\xspace}
\newcommand{\contSpace}[0]{\ensuremath{\mathcal{C}}\xspace}
\newcommand{\contVal}[0]{\ensuremath{c}\xspace}
\newcommand{\contextSpace}[0]{\Xi}
\newcommand{\context}[0]{\xi}
\newcommand{\calM}[0]{\ensuremath{\mathcal{M}}\xspace}

\newcommand{\costFn}[0]{R}

\newcommand{\regionpolicy}[0]{\ensuremath{\pi_{\discVal}}}
\newcommand{\costtogo}[0]{\ensuremath{J^{0:K}_{\discVal}}}

\newcommand{\bfp}{\ensuremath{\mathbf{p}}\xspace}

\newtheorem*{theorem-non}{Theorem}

\newtheorem*{lemm-non}{Lemma}

\title{Hybrid Planning for Dynamic \\ Multimodal Stochastic Shortest Paths}

\author{
Shushman Choudhury \\
Department of Computer Science\\
Stanford University\\
\texttt{shushman@stanford.edu} \\
\And
Mykel J. Kochenderfer \\
Department of Aeronautics and Astronautics\\
Stanford University\\
\texttt{mykel@stanford.edu}\\
}

\begin{document}

\maketitle

\begin{abstract}

Sequential decision problems in applications such as manipulation in warehouses,
multi-step meal preparation, and routing in autonomous vehicle networks often involve 
reasoning about uncertainty, planning over discrete modes as well as continuous states, and 
reacting to dynamic updates. 
To formalize
such problems generally, we introduce a class of Markov Decision Processes (MDPs)
called Dynamic Multimodal Stochastic Shortest Paths (DMSSPs).
Much of the work in these domains solves deterministic variants, which
can yield poor results when the uncertainty has downstream effects. 
We develop a Hybrid Stochastic Planning (HSP) algorithm, which uses domain-agnostic abstractions to efficiently unify
heuristic search for planning over discrete modes, approximate dynamic programming for stochastic planning over continuous states,
and hierarchical interleaved planning and execution.
In the domain of autonomous multimodal routing, 
HSP obtains significantly higher quality solutions than a state-of-the-art Upper Confidence Trees algorithm
and a two-level Receding Horizon Control algorithm.
\end{abstract}

\section{Introduction}
\label{sec:intro}

Consider the problem of a robot arm picking and arranging objects from a conveyor belt.
This simple example captures several challenges of sequential decision-making for robotics:
(i) the system state is a hybrid of discrete logical modes (is the robot holding an object or not)
and continuous robot state values (joint angles);
(ii) external information that constrains the inter-modal transitions (the object positions that define where they can be picked up)
is dynamically changing;
(iii) the objective is
to reach a goal (a given arrangement) with minimum cumulative trajectory cost, i.e. a stochastic shortest paths problem~\cite{bertsekas1991analysis},
which is challenging because the cost of a solution depends on
both the high-level sequence of objects grasped and the underlying motor control actions.
We define a class of Markov Decision Processes with the above properties, which we call the 
Dynamic Multimodal Stochastic Shortest Path (DMSSP) problem.
DMSSPs can represent many general decision-making problems of importance in robotics, such as task and motion
planning for mobile manipulation~\cite{wolfe2010combined,tan2003integrated} and autonomous multimodal routing~\cite{choudhury2019dynamic}.

Work on such robotics planning problems (hybrid state space, uncertainty, online information,
solution quality as the objective) 
have largely focused on efficiently solving deterministic variants,
delegating management of uncertainty to a low-level controller and replanning when it fails.
For a time-constrained setting, however, uncertainty in the dynamics may have significant downstream effects,
e.g. not reaching an object in time may invalidate the plan and greatly increase cost.
A framework that accounts for uncertainty in a high level plan will be more robust
to anticipating and avoiding downstream errors and temporal constraint violations.
We are motivated by designing such a framework while mitigating the inevitable increase in complexity due to considering uncertainty.

Existing planning algorithms for solving large MDPs, which do account for uncertainty,
would encounter several hurdles with DMSSPs. 
For offline MDP methods based on value or policy iteration, even with hierarchical decomposition~\cite{dietterich2000hierarchical}
and state-of-the-art function approximators~\cite{busoniu2010reinforcement}, it is typically infeasible to generate
a good policy over the entire space of external information.
Online MDP methods based on stochastic tree search
suffer from the large depth and branching factor for long-horizon search over continuous states and discrete modes.

Our work builds upon the idea that a principled composition of classical search-based planning, 
planning for MDPs, and hierarchical planning can \emph{reason over long horizons, explicitly account for the underlying uncertainty, 
and replan efficiently online}. 
% Through a domain-agnostic representation, we unite techniques from those fields
% in a way that we expect will achieve better
% quality solutions on DMSSPs than existing MDP methods.
We develop an algorithmic framework with three broad components: 
a global open-loop layer that plans for a sequence of discrete modes,
a local closed-loop layer that controls the agent under uncertainty through the modes, 
and hierarchical interleaved planning and execution to adapt to dynamic external information.
We expect the resulting approach to achieve better quality solutions on DMSSPs than existing MDP methods.

Our \textbf{contributions} are as follows: (i) We introduce and formulate the problem of Dynamic Multimodal SSPs. 
(ii) We design a Hybrid Stochastic Planning algorithm for decision-making in DMSSPs, which uses a careful choice of
domain-independent representations and abstractions to efficiently incorporate
heuristic search, approximate dynamic programming, and hierarchical interleaved planning and execution. 
(iii) We demonstrate how our approach outperforms two complementary baselines---a state-of-the-art Upper Confidence Trees algorithm and a 
two-level Receding Horizon Control algorithm---on real-time multimodal routing problems.

\textbf{Related Work Overview}

We provide a brief summary of the concepts we build upon.
Our formulation is based on Markov Decision Processes (MDPs)~\cite{puterman1994markov}
and Stochastic Shortest Paths~\cite{bertsekas1991analysis}, an undiscounted goal-directed negative-reward MDP. Our DMSSPs model share elements
with previously studied MDP models: arbitrarily modulated transition functions~\cite{yu2009arbitrarily},
stochastic shortest paths with online information~\cite{neu2010online}, and factored hybrid-space MDPs~\cite{kveton2006solving}.
Our HSP algorithm uses ideas from heuristic search~\cite{hart1968formal,pearl1985heuristics} and search-based planning for multi-step tasks~\cite{russell2003artificial,hoffmann2001ff}, approximate dynamic programming~\cite{bertsekas2005dynamic,busoniu2010reinforcement}, hierarchical planning for solving large MDPs~\cite{parr1998hierarchical,hauskrecht1998hierarchical,hengst2012hierarchical}, and interleaved planning and execution~\cite{nourbakhsh1997interleaving,lemai2004interleaving}.
A body of relevant previous work incorporates heuristic search and classical AI techniques in
algorithms for solving 
MDPs~\cite{mausam2007hybridized,kolobov2011heuristic,kolobov2012planning}.
Several works from the robotics planning community solve related problems using local information to inform global planning and trajectory optimization~\cite{atkeson1994using,choudhury2016regionally}
and explore
various aspects of combined (discrete) task and (continuous) motion planning~\cite{tan2003integrated,wolfe2010combined} such as planner-agnostic abstractions~\cite{srivastava2014combined},
stochastic shortest path formulations~\cite{srivastava2018anytime,kaelbling2013integrated}, and hierarchical
planning and execution~\cite{kaelbling2011hierarchical,kaelbling2016implicit}. Our optimization-based formulation is
similar to that for Logic-Geometric Programming~\cite{toussaint2015logic}.

% MDP; SSPs; factored mdps; logical planning; hybrid MDPs (see TAMP MDP); hierarchical planning (stoch and det); interleaving

\section{Dynamic Multimodal Stochastic Shortest Paths (DMSSPs)}
\label{sec:formulation}

A DMSSP is a discrete-time MDP, $\calM = (\calS, \calA, \contextSpace, T, R)$.
The \textbf{state space} is \emph{multimodal}, i.e. factored as $\calS \equiv \discSet \times \contSpace$. 
Each $\discVal \in \discSet = \{1,\ldots,{\absDiscSet}\}$ represents a discrete logical mode of the system and $\contSpace$ is the continuous state space of the decision-making agent.
The current system state is accordingly denoted as $s_t = \left(\discVal_t,\contVal_t\right)$.
The \textbf{action space} is $\calA \equiv \calA_{\discSet} \cup \calA_{\contSpace}$ where 
$\calA_{\discSet}$ is the set of discrete mode-switching actions (e.g. Pickup and Putdown) and $\calA_{\contSpace}$ is the set of control actions
for the agent.
The \textbf{context space} $\contextSpace$ is \emph{dynamic}.
At each time-step, the agent observes a discrete set of contexts $\{\context_t,\context_{t+1 \mid t},\context_{t+2 \mid t},\ldots,\context_{t + K \mid t}\}$. 
Here, $\context_t$ is the current context and $\context_{t+k \mid t}$ is the estimated context $k$ time-steps in the future. In general, $\context_{t+k \mid t} \neq \context_{t+k \mid t+1}$. The context is assumed to be generated by an exogenous process whose evolution equation, i.e. $\context_{t+1} = g(\context_t)$,
is complex and depends on a number of unobserved variables.
The current context set is compactly denoted as $\context_{t:t+K}$. The set cardinality $K$ is some domain-dependent prediction horizon, and the dimensionality of context is time-varying, i.e. $| \context_t |$ may differ from $|\context_{t+1 \mid t}|$. In our example, the context is the current and estimated future positions of all moving objects on the belt.

\begin{figure}[t]
    \begin{subfigure}{0.229\textwidth}
    \centering
        \fbox{\includegraphics[width=0.95\textwidth]{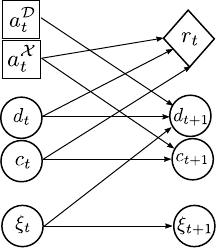}}
        \caption{}
        \label{fig:dmssp-diag}
    \end{subfigure}
    \begin{subfigure}{0.771\textwidth}
    \centering
        \fbox{\includegraphics[width=0.95\textwidth]{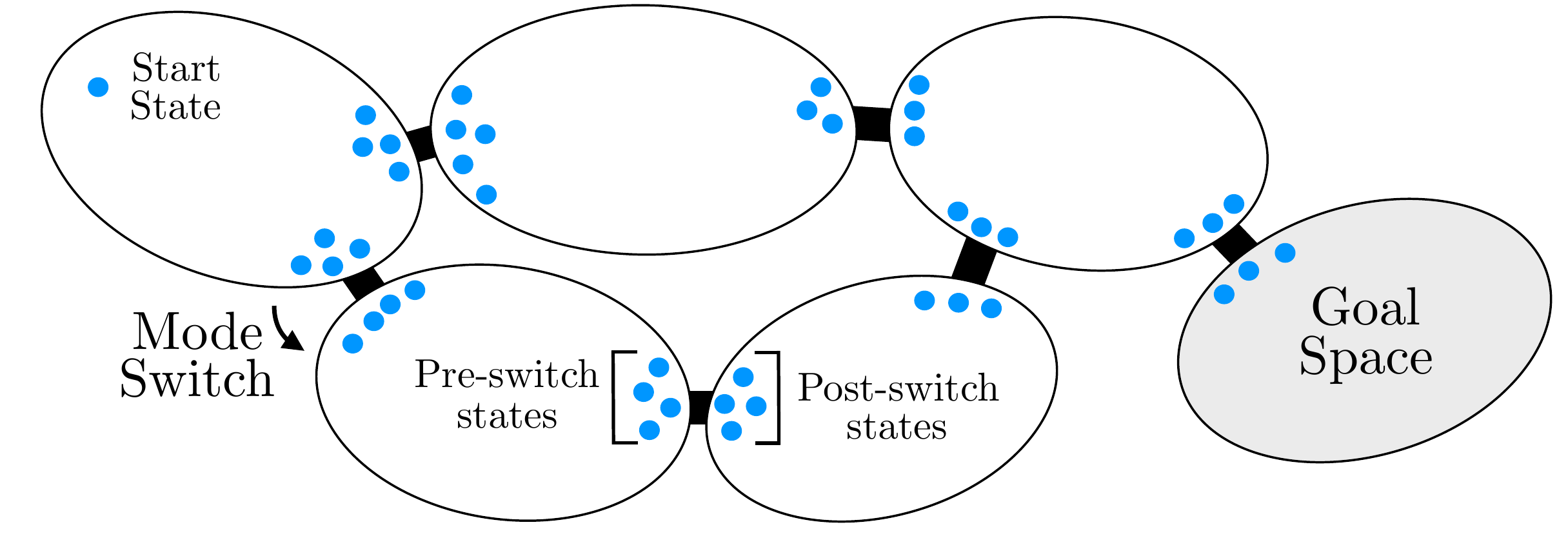}}
        \caption{}
        \label{fig:dmssp-viz}
    \end{subfigure}
    \caption{(\subref{fig:dmssp-diag}) The decision diagram for a DMSSP (notation from~\cref{sec:formulation}). (\subref{fig:dmssp-viz}) An abstract visualization of 
    a DMSSP. From a start state, the agent must reach
    a goal space. 
    To do so, it must choose a valid sequence of discrete logical modes and
    also control actions to navigate within the continuous subspaces corresponding to each mode.
    The mode switches can only happen at sampled transition states (border dots) that satisfy the 
    pre-switch conditions (or preconditions). Both the pre-switch and the post-switch states
    are constrained by the external context which is dynamically changing.
    The total cost of the solution is accumulated over the entire trajectory.}
    \label{fig:dmssps}
\end{figure}

The \textbf{transition function} $T$ can be factored as follows:

\begin{subequations}
    \begin{minipage}{0.49\textwidth}
    \begin{equation}
    \label{eq:transition-cont}
        T : \discSet \times \contSpace \times \calA_{\contSpace} \times \contSpace \rightarrow [0,1]
    \end{equation}
    \end{minipage}
    \begin{minipage}{0.49\textwidth}
    \begin{equation}
    \label{eq:transition-disc}
     T: \contextSpace \times \calS \times \calA_{\discSet} \rightarrow \calS
     \end{equation}
     \end{minipage}
\end{subequations}

The factored form efficiently encodes stochastic intra-modal physical dynamics in~\cref{eq:transition-cont} 
and context-dependent deterministic mode-switching rules in~\cref{eq:transition-disc}. 
We treat the discrete mode-switching as single time-step actions.
The grounded values of logical predicates (for mode-switching) are represented as states~\cite{srivastava2018anytime},
i.e. instances of $\calS$ and not just $\discSet$ in~\cref{eq:transition-disc}.
The current context $\context_t \in \contextSpace$ restricts the actual grounding of the logical predicates in the
preconditions and effects 
that define the mode switch rules.
Precondition states need to be reached for a mode switch to be feasible and effect states are 
obtained after the mode switch is completed.
\emph{In our running example, whether the end-effector
can satisfy the preconditions to grasp a moving object depends on the end-effector (state space) and the
object (context)}.
As in typical stochastic shortest paths~\cite{bertsekas1991analysis},
the problem is episodic and undiscounted. Also, the \textbf{reward function} is non-positive, so we will also use `cost' to represent negative rewards.
It is factored in terms of the intra-modal costs, i.e. $R : \discSet \times \contSpace \times \calA_{\contSpace} \times \contSpace \to \mathbb{R}^{-}$.
The decision diagram of a DMSSP is depicted in~\cref{fig:dmssp-diag}.

% \subsubsection*{Bridging Function}
% With some overloading of notation, we define the \emph{bridging function} $\beta$ as follows:
% \begin{equation}
%     \label{eq:bridging}
%     \begin{aligned}
%         \bridge_{\discVal \rightarrow \discVal_j}^{\context_t} &= \{(\contVal_1,\contVal_2) \mid \exists a \in \calA_{\discSet} \ \text{s.t.} \ T_{\discVal \rightarrow \discVal_j}^{\context_t} \left( (\discVal,\contVal_1), a, (\discVal_j,\contVal_2) \right) = 1 \}\\
%         \bridge_{\discVal \rightarrow \discVal_j}^{\context_t} (\contVal_1) &= \{ \contVal_2 \in \contSpace \mid (\contVal_1,\contVal_2) \in \bridge_{\discVal \rightarrow \discVal_j}^{\context_t} \}
%     \end{aligned}
% \end{equation}
% The bridging function essentially constrains the set of continuous states that can be reached after completing a mode switch from $\discVal$ to $\discVal_j$, given the state $\contVal_1$ from which the mode switch is attempted, and the context $\context_t$. Typically, $\bridge_{\discVal \rightarrow \discVal_j}^{\context_t} (\contVal_1)$ is a small fraction of $\contSpace$ and can be obtained analytically.

% \subsection{Solving a DMSSP}
% \label{sec:formulation-solving}

% The objective for solving any MDP is to obtain a policy which maps states to actions such that the resulting
% trajectories have minimum expected cost.
We explicitly write out a DMSSP as an online stochastic optimal control problem. Given a start state $s_0 = (\discVal_0,\contVal_0)$ and goal space $\calS_G = (\discVal_G,\contSpace_G \subseteq \contSpace)$, the overall problem is:
\begin{equation}
    \label{eq:problem}
    \begin{aligned}
    & \underset{\discVal_{1:H-1},a_{0:H-1}}{\mathrm{argmin}} 
    & & - \sum\nolimits_{t=0}^{H-1} \left[ R(\discVal_t,\contVal_t,a_t)\right] \ \ & \text{where} \ s_t = \left(\discVal_t, \contVal_t\right)\\
    % & \text{where}
    % & & s_t \equiv (\discVal_t,\contVal_t) \\
    & \text{subject to}
    & & \contVal_{t+1} \sim T(\discVal_t,\contVal_t,a_t) \ & \text{when} \ a_t \in \calA_{\contSpace} \ \text{(from \cref{eq:transition-cont})} \\
    & & & s_{t+1} =  T\left(\context_t,s_t,a_t \right) \ & \text{when} \ a_t \in \calA_{\discSet{}} \ \text{(from \cref{eq:transition-disc})} \\
    & & & s_H \in \calS_G & \text{reach goal space}
    \end{aligned}
\end{equation}
where $H$ is the number of steps taken.
From~\cref{fig:dmssp-viz},
a DMSSP planning algorithm has to choose a sequence of modes from start mode $\discVal_0$ to goal mode $\discVal_G$, 
and also control the agent to traverse through the corresponding subspaces.
Even the one-shot deterministic problem is a discrete-time Logic-Geometric Program~\cite{toussaint2015logic}, 
for which a globally optimal solution is infeasible. Theoretical optimality analysis requires several assumptions
and is not our focus (additional comments in appendix A).
We are interested in high quality solutions that are efficient.

% Our focus is to develop a solution framework for DMSSP that enjoys a balance of solution quality (compared to the best solution in hindsight) and scalability to large problem sizes. It is ill-posed to expect any theoretical guarantees of optimality in such settings. Since DMSSP is a class of problems, we seek an algorithmic framework that can accommodate a variety of subroutines, as is appropriate for the domain of interest. After introducing the general framework, we will discuss some instantiations of domain-dependent subroutines.

\section{Hybrid Stochastic Planning (HSP)}
\label{sec:hsp}

\begin{figure}[t]
    \centering
    \includegraphics{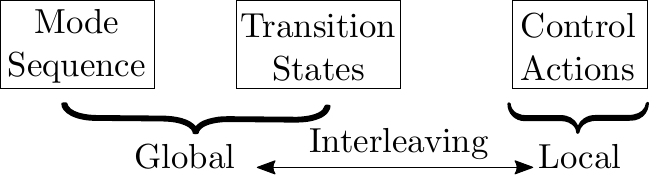}
    \caption{The global open-loop layer of HSP jointly decides the current valid mode sequence and transition states for the 
    mode switches. The local closed-loop layer controls the agent under uncertainty between entry and
    exit transition states of the regions. There is additional logic between the layers for (hierarchical) interleaving of planning and execution.}
    \label{fig:hsp-archi}
\end{figure}

The challenges with DMSSPs preclude directly applying existing techniques for
large MDPs, as mentioned in~\cref{sec:intro}. 
% The unbounded indeterminacy of the context 
% necessitates online replanning, while the bounded indeterminacy of the
% dynamics calls for contingent planning (c.f.~\cite{russell2003artificial}, chap. 11).
We develop a Hybrid Stochastic Planning framework that is particularly
suited for DMSSPs.
A global layer computes a sequence of mode switches and corresponding transition states;
planning open-loop enables efficiently searching over long-horizon mode sequences.
A local layer executes 
actions that control the agent within each mode; closed-loop planning provides some robustness to uncertainty.
Additional hierarchical logic for interleaving planning and execution reacts to the dynamically changing context at both global and local levels.
~\Cref{fig:hsp-archi} depicts a schematic of the HSP structure.  
The concepts we rely on have been studied extensively.
However, what we particularly contribute are the design choices in the overall framework (discussed subsequently);
heuristic search over mode sequences, decomposition to smaller MDPs, using cost-to-go functions, i.e. negative value functions as edge weight surrogates, pre-empting
the local controller.
These choices unite techniques from classical search-based planning, planning for MDPs, and hierarchical planning
in a principled manner.
% The following is a summary of the framework:
% \begin{itemize}
%     \item The global layer uses heuristic implicit tree search to jointly choose a sequence of mode switches 
%     and corresponding transition states that are valid \textit{as per the current context set}.
%     \item We define a smaller region-based MDP with carefully chosen terminal pseudo-costs. 
%     The \textit{corresponding cost-to-go function is used in the edge weight} for the global layer's tree search. 
%     The horizon-dependent closed-loop policy for the region's MDP controls the agent within the region while accounting for the temporal dynamics of the context set. 
%     \item The edges of the global layer tree are equivalent to states for the corresponding region's MDP. This equivalence allows seamless interleaving of planning (of paths on the tree) and execution (of the low-level actions for traversing the edges).
% \end{itemize}
\emph{Since DMSSP is a class of problems, we present a general algorithmic framework that can accommodate a variety of modular techniques and subroutines}.

\subsection{Global Open-loop Layer}
\label{sec:hsp-global-layer}

This layer repeatedly computes a high-level plan from the current state $s_t = (\discVal_t, \contVal_t)$ to a goal state $(\discVal_G, \contVal_G \in \contSpace_G)$.
The current high-level plan (computed at current time $t$) is denoted as follows:
\begin{equation}
\label{eq:current-plan}
\zeta_t = \left(\discVal^{0},\contVal^0\right) \xrightarrow{\pi_{\discVal^0}} \left(\discVal^{0},\contVal_{p}^{0}\right) \xrightarrow{a \in \calA_{\discSet}} \left(\discVal^{1},\contVal_{e}^{1}\right) \xrightarrow{\pi_{\discVal^1}} \left(\discVal^1,\contVal_p^{1}\right) \xrightarrow{a \in \calA_{\discSet}} \left(\discVal^2, \contVal_e^{2}\right) \rightsquigarrow \left(\discVal_G,\contVal_G\right)
\end{equation}
where $(\discVal^0,\contVal^0) = (\discVal_t,\contVal_t)$ and the sequence of chosen modes is $\discVal^0 \rightarrow \discVal^1 \rightarrow \ldots \rightarrow \discVal_G$. For each planned 
mode switch $\discVal^{n} \rightarrow \discVal^{n+1}$, the sampled precondition is $\contVal_p^{n}$ and effect is $\contVal_e^{n+1}$.
For each mode, $\pi_{\discVal^n}$ is a closed-loop policy that controls the agent through the region
from the effect after the previous switch $\contVal_e^n$ to the precondition before the next switch $\contVal_p^{n}$. \textsc{GlobalPlan} in~\Cref{alg:HSP-global} outlines the global layer, which runs a heuristic search~\cite{hart1968formal,russell2003artificial}
from the current state $s_t$ to compute
a sequence of modes and transition states towards the goal space $\calS_G$.
We briefly discuss the subroutines.

% Each time the layer is invoked, it runs an implicit tree search rooted at the current state $s_t$.
% The priority queue is based on the lexicographic ordering of $\text{key}(s) = \langle w(s) + \mathrm{heur}(s), w(s) \rangle$, where $w(s)$ is the cost-to-come on the current tree, and $\mathrm{heur}(s)$ is a heuristic estimate of the cost-to-go. 
% For each expanded node, $(\discVal_{\text{top}}, \contVal_{\text{top}})$, the set of valid next modes $\textsc{Next}(\discVal_{\text{top}})$ is generated.
% For each such mode $d^{\prime}$, the \textsc{SampleTransitions} subroutine generates $N$ candidate pairs of groundings for preconditions and effects $(\contVal_p^{\discVal_{\text{top}}}, \contVal_e^{\discVal^{\prime}})$(a higher $N$ increases expected solution quality at the expense of efficiency). It also generates a probability distribution over future timesteps $P_{1 : K}$ at which the
% context will allow the mode transition. 
% For each sampled pair, the post-transition effect state, $s^{\prime} = (\discVal^{\prime},\contVal_e^{\discVal^{\prime}})$, is inserted.
% Its cost-to-come $w^{\prime}$ is obtained by augmenting its predecessor's cost-to-come $w_{\text{top}}$ with the edge weight function. 
% A Predecessor pointer connects $s^{\prime}$ to $s_{\text{top}}$ and the grounded precondition, i.e. $\contVal_{p,i}^{\discVal_{\text{top}}}$.

For $\textsc{NextValidModes}$, we use the mode-switch rules and the current context set to identify possible next modes
from $\discVal_{\text{top}}$. The specific implementation is domain-dependent.
In our example, the robot can only execute Pickup on objects that will be reachable based on the current context set.
The \textsc{SampleTransition} method generates grounded precondition and effect states for
the proposed mode switch given the context set, using some domain-specific sampling procedure
(the number of samples $N$ is a parameter). It also generates $P_{1:K}$, a probability distribution over future time-steps at which the context $\context_{t+k}$ will satisfy the mode transition preconditions.
In our example, we could use work on sampling end-effector poses for grasping a given object~\cite{miller2003automatic}
at various steps along the object's expected future trajectory (defined by the context).

% takes the proposed mode switch $\discVal_{\text{top}} \rightarrow d^{\prime}$, a sample in the current mode $\contVal_{\text{top}}$, and $\context_{t:t+K}$ as input. It samples a context from the mode-switch context space $\context \sim \contextSpace_{\discVal_{\text{top}} \rightarrow d^{\prime}}$ and obtains the distribution $P_{1:K}$ over future time-steps where $\context$ will be true, based on the current context set $\context_{t:t+K}$. It then samples two bridge points $\{\contVal_i^{\discVal_{\text{top}}}, \contVal_i^{\discVal^{\prime}}\} \sim \beta^{\context}_{\discVal_{\text{top}} \rightarrow d^{\prime}}$. The additional argument $\contVal_{\text{top}}$ may be useful to filter out bridge points completely unreachable from $\contVal_{\text{top}}$.

For \textsc{EdgeWeight}, consider its role in the overall solution.
Each call to \textsc{GlobalPlan} returns a sequence of modes and transition states.
\textit{Subsequent execution by the local layer can only optimize the trajectory locally}.
The edge weight function should therefore be a good surrogate of the expected cost to have any hope of a good overall solution.
The cumulative intra-modal dynamics cost depends on the actual trajectory that will eventually be executed by the agent in that mode. Therefore, as a surrogate,
we use the \emph{cost-to-go function} of the policy for a local MDP corresponding to the mode, where the 
local MDP state is encoded in the edge (details explained in~\cref{sec:hsp-local}):
\begin{equation}
    \label{eq:edge-weight}
    \textsc{EdgeWeight}(P_{1:K}, \langle \discVal_{\text{top}}, \contVal_{\text{top}} \rangle \rightarrow
    \langle \discVal^{\prime} ,\contVal_{p} \rangle)
    = \sum\nolimits_{k=1}^{K} P(k) J^k_{\discVal_{\text{top}}}( \langle \contVal_{\text{top}}, \contVal_{p} \rangle) , \ \ k = 1 \ldots K.
\end{equation}
Here, $J^{0:K}_{\discVal_{\text{top}}}$ is the horizon-dependent cost-to-go function for the local MDP of mode $\discVal_{\text{top}}$.
\emph{We have presented the most general search formulation for the global layer}. It can use many of the speedup techniques
for heuristic search-based planning~\cite{hoffmann2001ff,helmert2009landmarks} to improve efficiency.
The heuristic function ($\mathrm{heur}$) in general is defined on $\discSet \times \discSet$; additional comments on the heuristic are in appendix B.

\begin{algorithm}[t]
\caption{HSP - Global and Local Layers}
\begin{algorithmic}[1]
\Procedure{GlobalPlan}{$s_t = \left(\discVal_t,\contVal_t\right), \calS_G = \left(\discVal_G,\contSpace_G \in \contSpace\right),\context_{t:t+K}, N$}
\State $Q \gets \text{PriorityQueue}\left(\{s_t,\text{key}(s_t)\}\right)$ \Comment{$\text{key}(s_t) = \langle \mathrm{heur}(\discVal_t,\discVal_G), 0 \rangle$}
\While{$Q$ not empty}
    \State $s_{\text{top}}, \langle (w_{\text{top}},\cdot \rangle \gets \mathrm{pop}(Q)$ \Comment{$s_{\text{top}} = (\discVal_{\text{top}},\contVal_{\text{top}})$}
    \State \algorithmicif \ $s_{\text{top}} \in \calS_G$ \ \algorithmicthen \ \textbf{return} $\zeta_t \gets \textsc{Path}(s_\text{top})$ \Comment{Backtrace with Predecessor}
    %\State Generate $\textsc{NextValidModes}(\discVal_{\mathrm{top}})$
    % \State $\textsc{NextValidModes} = \{ \discVal' \in \discSet \mid 
    % \exists \contVal_1,\contVal_2 \in \contSpace, \context \in \context_{t:t+K}, a \in \calA_{\discSet}, \ \text{s.t.} \ T(\context, (\discVal_{\text{top}},\contVal_1), a) = (\discVal',\contVal_2)  \}$ \label{lst:HSP-next}
    \For{$\discVal^{\prime} \in \textsc{NextValidModes}(\discVal_{\mathrm{top}},\context_{t:t+K})$}
      \For{$i = 1 \text{ to } N$} \Comment{Number of samples (parameter)}
        \State $\{\contVal_{p}, \contVal_{e}, P_{1:K}\} \gets 
        \textsc{SampleTransition}(\context_{t:t+K}, \discVal_{\text{top}},\discVal^{\prime})$ \label{line:HSP-sampletrans}
        \State $w^{\prime} \gets w_{\text{top}} + \textsc{EdgeWeight}(P_{1:K}, \langle \discVal_{\text{top}}, \contVal_{\text{top}} \rangle \rightarrow
      \langle \discVal^{\prime} ,\contVal_{p} \rangle)$ \label{lst:HSP-edgewt}
        \State $s^{\prime} \gets (\discVal^{\prime},\contVal_{e})$ \Comment{$\text{key}(s^{\prime}) \gets \langle w^{\prime}+\mathrm{heur}(\discVal^{\prime}, \discVal_G), w^{\prime} \rangle$}
        \State $\mathrm{insert}(Q,\{s^{\prime}, \text{key}(s^{\prime})\}), \ \ \text{Predecessor}(s^{\prime}) \gets (s_{\text{top}}, \contVal_{p})$
      \EndFor
    \EndFor
\EndWhile
\State \textbf{return} $\zeta_t = (\discVal^0, \contVal^0) \rightarrow (\discVal^0,\contVal_p^0) \rightarrow (\discVal^1, \contVal_e^1) \rightsquigarrow (\discVal_G, \contVal_G \in \contSpace_G)$
\EndProcedure
\Statex
\Procedure{LocalPreprocessing}{$\discVal,K$} \Comment{$\discVal \in \discSet$}
% \State $\mathcal{S}_{\regionmdp} \gets \contSpace \times \contSpace$, $\regionmdp \equiv (\mathcal{S}_{\regionmdp},\calA_{\contSpace},
%     T_{\regionmdp},C_{\regionmdp})$
\State Compute $\phi_{\discVal}$ from $(\contSpace \times \contSpace,\calA_{\contSpace}, T_{\discVal},R)$ \Comment{\Cref{eq:terminalcost}}
\State $J_{\discVal}^{0:K}, \ \bar{J}_{\discVal}^{1:K} \gets \textsc{FiniteHorizonValueIteration}(\discVal,\phi_{\discVal},K)$ \Comment{\Cref{eq:finite-vi}}
\State \textbf{return} $J_{\discVal}^{0:K}, \ \bar{J}_{\discVal}^{1:K}$ \Comment{$\bar{J}_{\discVal}^{1:K}$ used in~\cref{eq:interrupt-logic}}
\EndProcedure
\end{algorithmic}
\label{alg:HSP-global}
\end{algorithm}

\subsection{Local Closed-loop Layer}
\label{sec:hsp-local}

The local closed-loop layer in HSP controls the agent in its current mode up to the chosen transition state for the next switch. 
The layer is `local' because it is only provided information about the currently executing step of the current global plan $\zeta_t$, i.e. $(\discVal^{0},\contVal^0) \rightarrow (\discVal^{0},\contVal_{p}^0)$, where $\discVal^0$ is the current mode of the agent. For each mode $\discVal$, 
we define a local target-directed MDP $(\contSpace \times \contSpace, \calA_{\contSpace}, T_{\discVal}, R_{\discVal})$, where the first argument of the state
is the current position, and the second is the target. The control space is $\calA_{\contSpace}$. The $T_{\discVal}$ and $R_{\discVal}$ functions are derived from $T$ and $R$, i.e.
\begin{equation}
    \label{eq:region-trans-cost}
    \begin{aligned}
        T_{\discVal} (\langle \contVal, \contVal_g \rangle, a, \langle \contVal^{\prime} , \contVal_g \rangle) = T (\discVal, \contVal, a, \contVal^{\prime}) \ \ ; \ \ 
        R_{\discVal} (\langle \contVal, \contVal_g \rangle, a, \langle \contVal^{\prime} , \contVal_g \rangle) = R(\discVal, \contVal, a, \contVal^{\prime})
        % \\
        % R_{\regionmdp} (\langle \contVal_1, \contVal_2 \rangle, a, \langle \contVal_1^{\prime} , \contVal_2^{\prime} \rangle) &= R (\discVal, \contVal_1, a, \contVal_1^{\prime}) \text{ if }  \contVal_2 = \contVal_2^{\prime},\ 0 \text{ if } \contVal_2 \neq \contVal_2^{\prime}.
    \end{aligned}
\end{equation}
The Cartesian self-product $\contSpace \times \contSpace$ is required in general because the policy must be able to control the agent between any two states in $\contSpace$.
However, for many spaces, a state $\langle \contVal_1, \contVal_2 \rangle$ can be encoded with relative state $\contVal_1 \circ \contVal_2 \in \contSpace$, where $\circ$
is a difference operator,
and the target is always the `origin' $\contVal \circ \contVal$, i.e. the zero state.
\emph{Many modes typically share dynamics, so the same policy can be reused}~\cite{hauskrecht1998hierarchical}.
For example, in our setting, the robot dynamics effectively depend only on if an object is currently grasped or not, which can be encoded with two modal dynamics functions.

\textbf{Finite-Horizon Value Iteration}\\
The local closed-loop policy has a dual role, controlling the agent with low cost within the mode to the transition state $\contVal_p$ chosen by the global layer
and satisfying the temporal constraints of the context for the next mode switch ($P_{1:K}$). 
If it is too slow, the mode transition may fail, affecting the overall solution. 
On the other hand, controlling the agent as quickly as possible may be highly sub-optimal. To model this tradeoff, we use finite-horizon value iteration to obtain a horizon-dependent policy~\cite{bertsekas2005dynamic}.
The finite-horizon value iteration requires a horizon limit (we use the context horizon $K$) and a terminal cost $J^0_{\discVal}$. For all local MDP states $\hat{s} \in \contSpace \times \contSpace$, we compute for $k = 1 \ldots K$,
\begin{equation}
    \label{eq:finite-vi}
    J^k_{\discVal}(\hat{s}) = \min_{a \in \calA_{\contSpace}} J^k_{\discVal}(\hat{s},a) = \min_{a \in \calA_{\contSpace}} \sum\nolimits_{\hat{s}^{\prime} \in \contSpace \times \contSpace}
    T_{\discVal}(\hat{s},a,\hat{s}^{\prime})\left[ - R_{\discVal}(\hat{s},a,\hat{s}^{\prime}) + J^{k-1}_{\discVal}(\hat{s}^{\prime}) \right]
\end{equation}
where $J_{\discVal}^k$ represents both cost-to-go and action-cost-to-go (Q function), with overloading (negative reward i.e. positive cost). The full cost-to-go function, compactly denoted as $\costtogo$, can be used in~\cref{eq:edge-weight}. 
\emph{As in the global layer, we require only a general framework for value iteration; any local or global approximation scheme~\cite{busoniu2010reinforcement} and other approximate dynamic programming~\cite{bertsekas2005dynamic} techniques could be used.}
The regional closed-loop policy $\pi_{\discVal}$, invoked during execution online, is based on $\costtogo$ obtained offline. For the current DMSSP state $(\discVal_t, \contVal_t)$, context horizon distribution $P_{1:K}$, and transition state $\contVal_p$ (provided by the global layer), the control action chosen locally for $\hat{s} = \langle \contVal_t, \contVal_p \rangle$ is
\begin{equation}
    \label{eq:region-policy}
    a_t = \pi_{\discVal_t}(\hat{s}) = \underset{a \in \calA_{\contSpace}}{\mathrm{argmin}} \ J_{\discVal_t}^{0:\bar{K}}(\hat{s}) = \underset{a \in \calA_{\contSpace}}{\mathrm{argmin}} \sum\nolimits_{k=1}^{K} P(k) J^k_{\discVal_t}(\hat{s},a) \ .
\end{equation}

\textbf{Terminal Pseudo-Cost}\\
% The terminal cost function $J^0_{\discVal}$ guides the behavior of $J^{0:\bar{K}}_{\discVal}$. \emph{Successful states} for $\regionmdp$ are those where the current state is near the current target, i.e. $\regionalTerminal = \{ \langle \contVal_1, \contVal_2 \rangle \in \calS_{\regionmdp} \mid \lvert \contVal_1 - \contVal_2 \rvert \leq \epsilon_{\regionmdp} \}$ for some domain-dependent metric and threshold. 
To incentivize the closed-loop policy to reach the target, we need a terminal penalty for 
states where the target is not reached at horizon 0, 
i.e. $\{ \langle \contVal, \contVal_g \rangle \in \contSpace \times \contSpace \mid \lVert \contVal - \contVal_g \rVert > \epsilon_{\discVal}\}$,
for some domain-dependent distance metric.
We need to be careful while choosing the terminal penalty or pseudo-cost due to the sub-optimality of hierarchical MDP planning~\cite{dietterich2000hierarchical,hengst2012hierarchical}.
The penalty is not in the true cost function in~\cref{eq:problem}, so the higher it is set, the poorer is $\costtogo$ as a surrogate of the true cost, and the more (locally) sub-optimal is $\pi_{\discVal}$ as a controller. An insufficiently high penalty, on the other hand, may lead to $\pi_{\discVal}$ choosing lower-cost actions at the risk of being unable to reach the target within the context horizon. \emph{Consequently, the attempted mode switch may fail, forcing HSP to recompute a different mode sequence, leading to much poorer solutions (downstream effects of uncertainty)}. 

We set the penalty $\phi_{\discVal}$ as the maximum cost of any $K$-length action sequence within the mode, i.e.
\begin{equation}
    \label{eq:terminalcost}
    \phi_{\discVal} = \max_{\mathbf{a} \in \calA_{\contSpace}^K, \ \contVal \ \in \ \contSpace} \ - \sum\nolimits_{k=1}^{K} R(\discVal, \contVal,\mathbf{a}_{k}),
\end{equation}
The pseudo-cost is the smallest penalty that prioritizes the mode sequence. The $\phi_{\discVal}$ value can be computed offline and then used in the finite-horizon value iteration.
Further comments on the pseudo-cost and horizon limit are made in appendices A and C).
\textsc{LocalPreprocessing} in~\cref{alg:HSP-global} outlines the local layer; the policies obtained from pre-processing are used online.

An implicit assumption of ours
is that the MDP for the agent dynamics can be solved reasonably. This assumption is not always valid, e.g. a complex
underactuated system or an articulated manipulator.
However, for
many practical systems, framing and solving the control problem with MDPs has been successful~\cite{duan2016benchmarking}, and finite-horizon versions 
of those controllers could also be used here.

\subsection{Hierarchical Interleaved Planning and Execution}
\label{sec:hsp-interleaving}

Interleaving planning and execution is an important property for real-time decision-making. 
Our HSP framework uses the \emph{simplification} approach to interleaving~\cite{nourbakhsh1997interleaving}. 
The global layer simplifies the underlying intra-modal control problem by determinization and planning over multiple timesteps,
and computes a solution in this simplified space. The local layer executes the plan provided by the global layer. 
% Sub-goaling with partial plans is suitable only when the fringe has enough information to evaluate progress toward goal achievement. In DMSSPs, whether the next mode represents progress towards the goal depends in turn on the goal being reachable from it. This motivates using simplification. 
We discuss two aspects of our interleaving, each occurring at one of the levels, either global replanning or local pre-emption.
The full HSP framework is outlined in~\cref{alg:HSP-full}.

\begin{figure*}
    \begin{subfigure}{0.47\textwidth}
    \centering
        \includegraphics[width=0.8\textwidth]{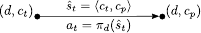}
        \caption{}
        \label{fig:interleaving-pre}
    \end{subfigure}
    \begin{subfigure}{0.53\textwidth}
    \centering
        \includegraphics[width=0.8\textwidth]{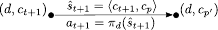}
        \caption{}
        \label{fig:interleaving-post}
    \end{subfigure}
    \caption{Our framework efficiently interleaves planning and execution. From time $t$ (\subref{fig:interleaving-pre}) to $t+1$ (\subref{fig:interleaving-post}), there is a change in the next mode transition state ($c_p \rightarrow c_{p'}$) due to the global layer. 
    This update at the planning level is immediately reflected at the execution level 
    by the change in the state of the local MDP ($\hat{s} \rightarrow \hat{s}_{t+1}$) and the
    corresponding action ($a_t \rightarrow a_{t+1}$) chosen by the local policy $\pi_d$.}
    \label{fig:interleaving}
\end{figure*}

\textbf{Global Replanning}\\
HSP uses a combination of \emph{event-driven} and \emph{periodic} replanning~\cite{church1992analysis}. 
The two events that trigger replans asynchronously are closed-loop pre-emption and failed mode-switch attempts. 
The current global plan is then invalidated and a new plan must be generated before execution can resume. 
With periodic replanning, the global layer computes a new global plan from the current state synchronously in the background, 
while the local layer executes the current plan. The duration of the replanning period is domain-dependent; but
a domain-agnostic strategy is to replan immediately after the previous plan has finished computing.
This updates the local layer's next target every $\Delta T$ time-steps, where $\Delta T > 1$ is the ratio between closed-loop and open-loop frequencies. \emph{Each time a global plan has been recomputed by the global layer, 
the next target of the local policy $\pi_{\discVal}$ is immediately updated to the (potentially) new transition state
for the next switch} (\cref{fig:interleaving} illustrates this).

\textbf{Local Pre-emption}\\
Updates to $\context_{t:t+K}$ can make it difficult to reach the chosen mode transition state in time (e.g. a target object
suddenly speeds up).
The periodic global replanning does account for this.
However, the latency is $\Delta T$ time-steps. 
We have additional closed-loop pre-emption logic to reason about the next chosen mode switch at the higher frequency of the local layer. For each local MDP, 
we compute and store (during pre-processing) the worst cost-to-go for any state from each horizon value, i.e.
$\bar{J}_{\discVal}(k) = \max_{\hat{s},a} J^k_{\discVal}(\hat{s},a), \ k=1\ldots K$. During the online execution of the local layer, if the agent is at a state 
from where reaching the goal in the remaining horizon is sufficiently unlikely, i.e.
\begin{equation}
    \label{eq:interrupt-logic}
    \text{if} \ \sum\nolimits_{k=1}^{K} P(k) J^k_{\discVal}(\hat{s}) > \beta_{\discVal} \cdot \sum\nolimits_{k=1}^{K} P(k) \bar{J}_{\discVal}(k) \ \text{then} \ a_t \gets \text{ Closed-loop Pre-emption}
\end{equation}
where $\beta_{\discVal} \in [0,1]$ is a risk parameter. The lower $\beta_{\discVal}$ is set, the lower the risk we are willing to take that the agent can reach the next transition state in time. \emph{In addition to a higher frequency for reasoning about pre-emption, this logic provides a modifiable risk aversion}.
%A closed-loop interrupt transfers control to the global layer and triggers an asynchronous replan.

\begin{algorithm}[t]
\caption{HSP - Full Framework (uses~\cref{alg:HSP-global})}
\begin{algorithmic}[1]
\Require $s_0 = (\discVal_0,\contVal_0), \calS_G = (\discVal_G,\contSpace_G \subseteq \contSpace), \context_{0:K}, (K,N,\{\beta_{\discVal}\}) $
\Statex $J_{\discVal}^{0:\bar{K}}, \ \bar{J}_{\discVal}^{1:K} \gets \textsc{LocalPreprocessing}(\discVal,K) \ \forall \ \discVal \in \discSet$ \Comment{Pre-processing}
\State $t \gets 0, \ \mathrm{plan} \gets true, \ \mathrm{lpt} \gets 0, \ s_t \gets s_0, \ \context_{t:t+K} \gets \context_{0:K}$
\Repeat
\If{$\mathrm{plan} = true$}
\State $\zeta_t = \textsc{GlobalPlan}(s_t, \calS_G, \context_{t:t+K}, N)$ \Comment{Global layer}
\State $\mathrm{lpt} \gets t, \ \mathrm{plan} \gets false$ \Comment{Reset plan flags}
\EndIf
\State $(\discVal_t,\contVal_t) \gets s_t, \ \contVal_{p} \gets \zeta_t[0]$ \Comment{Next transition (precondition)}
\If{$\contVal_t \approx \contVal_p$} \Comment{At transition state}
\State $(\discVal',\contVal_e) \gets \zeta_t[1]$
\State $a_t \gets a \in \calA_{\discSet} \ \text{s.t.} \ T(\context_t, s_t, a) = (\discVal', \contVal_e)$ \Comment{Mode-switch action}
\Else
\State \algorithmicif \ $\hat{s} = \langle \contVal_t, \contVal_p \rangle$ satisfies~\cref{eq:interrupt-logic} \ \algorithmicthen \ $a_t \gets \text{Pre-emption}$
\State \algorithmicelse \ $a_t \gets {\mathrm{argmin}}_{{a \in \calA_{\contSpace}}} \ J_{\discVal_t}^{0:\bar{K}}(\hat{s} = \langle \contVal_t, \contVal_p \rangle)$ \Comment{Local layer, \cref{eq:region-policy}}
\EndIf
\State $s_{t+1},\context_{t+1:t+1+K} \sim \mathrm{Environment}(s_t,a_t), \ t \gets t+1$ \Comment{Observe World}
\State \algorithmicif \ $\text{Pre-emption or failed switch or} \ t - \mathrm{lpt} \geq \Delta T$ \ \algorithmicthen \ $\mathrm{plan} \gets true$ \Comment{Replanning}
\Until{$s_t \in \calS_G$}
\end{algorithmic}
\label{alg:HSP-full}
\end{algorithm}

\section{Experiments: Multimodal Routing Domain}
\label{sec:experiments}

We use a different domain than our running example; the recently introduced Dynamic Real-time Multimodal
Routing (DREAMR) problem~\cite{choudhury2019dynamic}. We omit an elaborate description of the domain (see appendix E for more).
\emph{The DREAMR problem requires planning and executing routes
under uncertainty for an autonomous agent that can use multiple modes of transportation in a dynamic
transit vehicle network.} 
There are two discrete modes in the problem, \textsc{Move} for when the agent
moves by itself and \textsc{Ride} for when the agent uses transport. The continuous state is the agent's position
and velocity. The mode-switching actions
are \textsc{Board} and \textsc{Alight}, which switch \textsc{Move} to \textsc{Ride} and vice versa respectively. The noisy agent control
actions are for acceleration in each direction.
The agent is penalized for energy expended due to movement and waiting in place, and total elapsed time.
The transit vehicle routes are the contextual information; at any time, the current context set
comprises the current position and estimated future route (as a sequence of waypoints) of each active vehicle.
\emph{The estimated time of arrival (ETA) for each subsequent waypoint is subjected with some probability to a bounded two-sided deviation at each
timestep}. Mode switches can only be made at transit waypoints. Additionally, the agent can only \textsc{Board} a vehicle if it is sufficiently close
to the waypoint at the same time as the vehicle, and sufficiently slow.
The dimensionality of the context space increases
with the number of active route waypoints, and so is highly dynamic and very large (in the thousands).
Though there are technically only two modes, in practice, the number of valid mode sequences to the goal is exponential
in the number of transit vehicle routes.

\subsection{Baselines: Upper Confidence Trees and Receding Horizon Control}
\label{sec:experiments-baseline}

We use two \emph{complementary baselines} to evaluate the benefit of our solution. 
First, a (domain-specific) two-level Receding Horizon Control (RHC) method which repeatedly solves a deterministic problem.
It uses graph search for planning routes, and non-linear receding horizon control trajectory optimization
for executing them.
The second is based on Upper Confidence Trees (UCT)~\cite{kocsis2006bandit}, a general
online MDP planning algorithm, with two enhancements: (i) techniques from PROST~\cite{keller2012prost}, a state-of-the-art UCT-based probabilistic planner~\cite{vallati20152014}; 
(ii) double progressive widening~\cite{couetoux2011continuous} which artificially limits the branching factor and is more suitable for a continuous state space.
% We do expect HSP to outperform UCT on DREAMR as the latter is more general purpose while the former exploits the structure of DMSSPs;
% our evaluation goal is to investigate the relative performance. 
We additionally assist the UCT baseline: (i) the value function estimates and Q-value initializations are informed by $\costtogo$
(ii) the tree depth is set to the same horizon limit $K$ as the HSP local layer (iii) many trials are run to compute
good estimates for each action.

\begin{figure*}
\centering
\begin{minipage}[c]{0.55\textwidth}
\centering
\includegraphics[width=\textwidth]{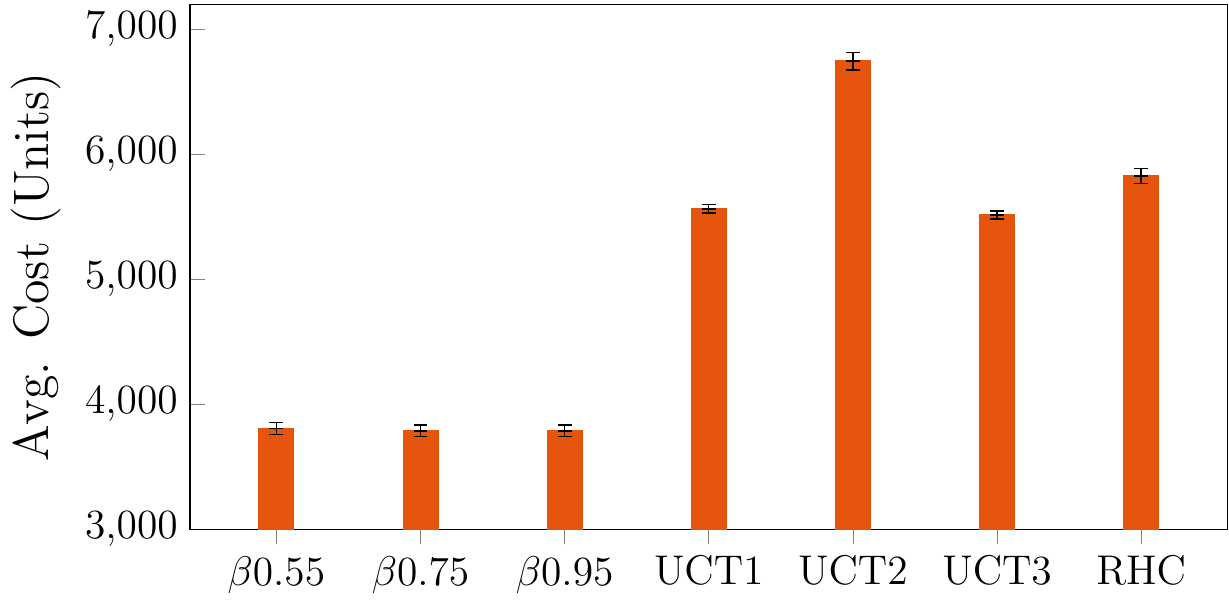}
\caption{Average cost (1000 problems) for HSP (three different values of the $\beta$ risk parameter), three variants of UCT, and RHC, with standard error bars. Appendix E.3 has results on two other sets of episodes.}
\label{fig:cost-time}
\end{minipage}
\hfill
\begin{minipage}[c]{0.4\textwidth}
\centering
\captionsetup{type=table} %% tell latex to change to table
\begin{tabular}{@{}rr@{}}
  \toprule
  Algorithm & Avg. Switches\\
  \midrule
  $\beta0.55$ (HSP) & 3.435 \\
  $\beta0.75$ (HSP) & 3.422 \\
  $\beta0.95$ (HSP) & 3.422 \\
  $\mathrm{UCT1}$ & 0.0243\\
  $\mathrm{UCT2}$ & 0.0176\\
  $\mathrm{UCT3}$ & 0.0154\\
  $\mathrm{RHC}$ & 3.508\\
  \bottomrule
  \end{tabular}
\caption{HSP (top three) chooses more mode switches than UCT and better mode switches than RHC, which explains its superior peformance in~\cref{fig:cost-time}.  }
\label{tab:avg-switches}
\end{minipage}
\end{figure*}

\subsection{Results}
\label{sec:experiments-results}

We used the POMDPs.jl framework~\cite{egorov2017pomdps} in Julia (additional details in appendix D and attached code\footnotemark).
\footnotetext{The Julia code is at \url{https://github.com/sisl/CMSSPs}}For HSP,
we chose three values for the $\beta \in [0,1]$ risk parameter for attempting time-constrained risk parameters: $0.55$, $0.75$ and $0.95$
(for lower values, it is overly risk-averse and rejects most transit connections).
For UCT, we chose three different combinations of 
the number of virtual rollouts at a new tree node and the UCB exploration constant.
The RHC does not have an important tuning parameter.
We used the same large-scale problem scenarios as the original DREAMR paper (see appendix E for more details and parameter values).
We evaluated each algorithm instantiation on $1000$ simulated episodes.
~\Cref{fig:cost-time} shows the average cumulative trajectory cost; both 
UCT and RHC produce poorer quality trajectories than HSP.
~\Cref{tab:avg-switches} displays the average number of
executed mode switches. 
\emph{Appendix E.3 has average computation times, and performance results on two additional sets of episodes with varying problem parameters}.
HSP has low sensitivity to values of $\beta > 0.5$ (at least on this domain), which is a useful property.
Multiple factors explain the relative performance gap between HSP and the baselines.
For UCT, compared to HSP, it uses far fewer modes of transportation. 
For most episodes, UCT controls the agent directly to
the goal without any mode switches. As a result, it incurs a far higher overall cost
for energy expended due to movement.
In general, a tree-based method such as
UCT requires a thorough search over the actions with very deep
lookahead to even possibly consider taking transportation, as we identified earlier.
On the other hand, RHC uses about the same number of mode switches as HSP, due to the similar
lookahead in its global layer enabled by deterministic planning. 
However, it uses a nominal edge weight for
the graph search, which is a poorer surrogate than the cost-to-go function,
and makes the choice of mode switches poorer.
It also uses receding horizon control rather than closed-loop
control at the local level, which makes it more sub-optimal locally, especially for making time-constrained connections.

\section{Discussion}
\label{sec:discussion}

We introduced and formulated the Dynamic Multimodal Stochastic Shortest Path problem
for representing sequential decision-making problems for complex robotics settings.
Our Hybrid Stochastic Planning framework, through our choice of abstractions, is a principled way of incorporating techniques from
heuristic search, approximate MDP planning, and interleaving planning and execution.
HSP's performance on the real-time autonomous routing domain against the complementary baselines highlights our general motivation.
By explicitly using online long-horizon planning and accounting for the underlying uncertainty and its downstream effects, 
we can achieve good quality solutions.
\emph{Our key limitations} are the assumptions on the various subroutines and components,
e.g. the mode transition states can be sampled efficiently from the context, the MDP for the agent dynamics can be solved, 
the value function lookup is fast, and so on.
However, as we mentioned, there are still several domains of interest where these 
assumptions are quite reasonable and have been used effectively,
and our work would be applicable in all of them.
Future research includes more 
detailed theoretical analyses with problem assumptions, using an online stochastic planner at the local layer
to overcome the need for an offline phase, and empirical results on other problem domains.

% \bibliographystyle{unsrtnat}
% \bibliography{refs.bib}

% !TEX root = ./neurips_2019.tex
\newpage

\appendix

All references are from the original bibliography.

\section{Comments on HSP Optimality}
\label{sec:appendix-opt}

We briefly stated in~\cref{sec:formulation} how the properties of DMSSPs,
namely the online context updates and the hybrid (discrete and continuous) state space,
make any useful theoretical analysis very difficult. We provide additional comments and justification for that here
and analyze quantitatively the role of the terminal pseudo-cost from~\cref{eq:terminalcost}. This section
is intended more to highlight the issues with optimality analysis of HSP
rather than prove any particular results (which would require several modeling assumptions and is not the scope of this work).

In general, analysis of an online optimization algorithm is done with respect to the best solution in hindsight,
but even this comparison typically assumes a specific functional form for the online information. However, for DMSSPs
the form of the context is entirely domain-dependent (e.g. route information in DREAMR or object trajectories in dynamic
TAMP). Therefore, for our subsequent discussion we will assume full observability of the true context
at all future timesteps.

\subsection{Global Optimality}
\label{sec:appendix-opt-global}

The cost of a solution depends on the discrete sequence of mode switches chosen as well as the underlying control
actions. In a deterministic setting alone, for a general non-linear cost function (which is the case for DMSSPs),
this is a Mixed Integer Program (MIP), which is known NP-Complete. The presence of uncertainty in the control
outcome makes this an even more difficult Stochastic Mixed Integer Program (SMIP), which is out of the scope of this discussion.
Practical solutions for MIPs use heuristic methods based on combinatorial techniques like 
tabu search, hill climbing, simulated annealing, and others.

In our case, HSP's global open-loop layer uses an \emph{anytime search} method parameterized by the number of samples
$N$ for each considered mode switch (see~\cref{alg:HSP-global}). A higher value of $N$, i.e. a greater amount of computation time devoted to the 
open-loop planning will yield better quality solutions. Of course, the quality of these solutions
is in terms of the edge weight, which is a surrogate objective for the true cost that depends on the actual executed
trajectory. Therefore, even in the \emph{asymptotic} case, i.e. as $N \rightarrow \infty$, it appears that no guarantee
of global optimality can be made. We use the cost-to-go function of the local MDP state encoded in the edge as a good surrogate of the expected cost.

\subsection{Local Optimality}
\label{sec:appendix-opt-local}

In the multimodal setting of DMSSP, local optimality refers to optimality
within the chosen mode sequence, i.e. whether the HSP solution
has minimum expected cost
out of all the solutions constrained to follow that mode sequence.
Due to the continuous component of the state space for each mode,
the expected cost-to-go within a region depends on the
approximation error $\epsilon_{\discVal}$ of the value iteration method used to obtain
the local policy $\regionpolicy$. Additionally, the global open-loop
planning would have to cover the space of all possible sampled pre-conditions
for each mode switch.
Therefore, we will consider
the case when the agent's component of the state space, $\contSpace$ is 
discrete rather than continuous, and when the global layer considers every possible discrete precondition during planning. 
As $N \rightarrow \infty$ and $\epsilon_{\discVal} \rightarrow 0 \ \forall \discVal \in \discSet$,
this discrete case performance will be emulated.

\textbf{N.B.} The following discussion relies heavily on section 7.2 of~\citet{bertsekas2005dynamic}, which discusses
Stochastic Shortest Path Problems for the discrete state space case.

For simplicity, we assume the following properties for the agent state space $\contSpace$ and every local intra-modal MDP:

\begin{itemize}
\item $\contSpace = \{\contVal_0,\ldots,\contVal_n\}$ is a set of discrete states.
\item There is a difference operator $\circ$ such that $\contVal_1 \circ \contVal_2 \in \contSpace$. Furthermore,
$\contVal \circ \contVal = \contVal_0 \ \forall \contVal \in \contSpace$. We can transform
the cost function as~$\costFn_{\discVal}(\langle \contVal_1, \contVal_2 \rangle, a, \langle \contVal'_1, \contVal_2 \rangle) 
= \costFn_{\discVal}(\contVal_1 \circ \contVal_2, a, \contVal'_1 \circ \contVal_2)$
\item The state $\contVal_0$ is the \emph{cost-free absorbing state}. We also refer to this as the `zero' state or origin, for obvious reasons.
\item There is at least one \emph{proper policy} (\cite{bertsekas2005dynamic}, cf. Assumption 7.2.1 footnote), i.e. a stationary policy which has non-zero probability of reaching the zero state $\contVal_0$ after some number of stages $m$,
regardless of the initial state. This assumption is actually quite weak in practice.
\end{itemize}

Given these assumptions, the HSP local layer sub-task of reaching state $\contVal_e$ from a start state $\contVal_p$ 
is equivalent to the classical stochastic shortest paths problem of reaching the zero state $\contVal_0$ from $(\contVal_p \circ \contVal_e)$ with 
minimum expected cost.
DMSSPs additionally have a finite-horizon setting for $\discVal$ because of the temporal constraints induced
on mode transitions by the context set. For the fully observable context set, the distribution over future
timesteps $P_{1:K}$ collapses to an exact time horizon, say $k$, within which the agent must reach the terminal
state, in order to successfully make the mode switch.
By Proposition 7.2.1 (a) of~\citet{bertsekas2005dynamic}, \emph{for the finite-horizon case, 
the value iteration algorithm of~\cref{eq:finite-vi} yields the optimal stage-wise, i.e. horizon-dependent cost 
$\costtogo$ from every start state, where the terminal cost is given by $J^0$}.

Given the start state  $\contVal_e$ in the current mode, if the global layer samples 
every possible discrete precondition $\contVal_p$ for the next mode-switch out of the current region,
then every possible relative start state $\contVal_p \circ \contVal_e$ would be considered,
and $\contVal^{*}_{p} = \mathrm{argmin}_{\contVal_p \in \contSpace} J^k_d(\contVal_p \circ \contVal_e)$ would be
chosen, where $k$ is the known time interval for the future context permitting the mode switch.

Therefore, at least in the discrete case, our representation of the local layer's local MDP
allows us to inherit the optimality properties of SSP problems. However, due to the 
finite-horizon setting, we are only optimal with respect to the terminal cost $J_0$.
The terminal cost issue illustrates the potential conflict at the local layer between reaching the target 
with a low cost and reaching the target in time, and is a caveat
to any local optimality analysis of DMSSPs. 

For a given mode switch chosen by the global layer, the sequence of actions
with minimum expected cost may not have the lowest probability of reaching the chosen
transition point, i.e. reaching the zero relative state in time, which would
lead to globally poorer overall trajectories. In a deterministic setting, we could have 
constrained sets of
control actions guaranteed to reach the zero state. However, for the stochastic setting of DMSSPs,
we must instead consider the probability of the action sequence to reach the zero state. We attempt to
balance this with the terminal pseudo-cost of~\cref{eq:terminalcost}, which we analyze further here.

\subsubsection*{Terminal Pseudo-Cost Analysis}

We are considering the finite-horizon value iteration of~\cref{eq:finite-vi} with the terminal cost
of~\cref{eq:terminalcost}. For the discrete state space case that we are analyzing,
the terminal (at horizon 0) relative state corresponding to a target being reached is the `zero' state $\contVal_0$.
All other states represent the target not being reached and are assigned the terminal penalty from~\cref{eq:terminalcost}.
Therefore, the terminal cost function $J^{\discVal}_0$ is defined as follows:
\begin{equation}
    \label{eq:appndx-terminal}
    J_0(\contVal_0) = 0 \ ; \ \ \ \ J_0(\contVal \neq \contVal_0) = \phi_{\discVal}.
\end{equation}

Denote (as we did earlier) the policy derived from the optimal cost-to-go function $\costtogo$ as $\pi^{0:K}_{\discVal}$ (denoted $\pi$ hereafter compactness).
For any given relative state $\contVal \in \contSpace$, define the probability of
the `zero' state being reached from it by following $\pi$ after $k$ steps as
\begin{equation}
\label{eq:appndx-reach-prob}
    \rho_{\pi}(\contVal) = P\{\contVal_{t+k} = \contVal_0 \mid \contVal_t = \contVal , \pi \}
\end{equation}
for the current time-step $t$. Furthermore, denote the k-stage expected cost-to-go for $\pi$
\emph{with terminal cost $0$ for all states} as
\begin{equation}
\label{eq:appndx-notermcost}
    \hat{J}_{\discVal}^k(\contVal \mid \pi) = \sum_{\contVal^{\prime} \in \contSpace \times \contSpace} T_{\discVal}(\contVal,\pi(a),\contVal^{\prime})\left[ - R_{\discVal}(\contVal,\pi(a),\contVal^{\prime}) + J^{k-1}_{\discVal}(\contVal^{\prime} \mid \pi) \right], \ \ 
    J_{\discVal}^0 = 0 \ \forall \contVal \in \contSpace
\end{equation}
where, once again, negative reward is used to imply positive cost (the reward function is non-positive).
Using~\crefrange{eq:appndx-terminal}{eq:appndx-notermcost}, we can express the general k-stage cost-to-go for $\pi$ as
\begin{equation}
\label{eq:appndx-reach-cost}
\begin{aligned}
J_{\discVal}^k(\contVal \mid \pi) &=  \rho_{\pi}(\contVal) \cdot \hat{J}_{\discVal}^k(c \mid \pi) + [1 - \rho_{\pi}(\contVal)] \cdot (\hat{J}_{\discVal}^k(c \mid \pi) + \phi_{\discVal} ) \\
&= \hat{J}_{\discVal}^k(c \mid \pi) + [1 - \rho_{\pi}(\contVal)] \cdot \phi_{\discVal},
\end{aligned}
\end{equation}
where the left-hand term is the expected cost due to the trajectory and the right-hand term is the penalty weighted by the 
probability of failure to reach the target in time (i.e. in $k$ steps). 
If we express our finite horizon value iteration as a finite horizon policy iteration (using the fact that a policy
can be extracted from a value function), then our corresponding policy search is
\begin{equation}
\label{eq:appndx-policy-iter}
\pi^{*} = \mathrm{argmin}_{\pi \in \Pi} \ \hat{J}_{\discVal}^k(c \mid \pi) + [1 - \rho_{\pi}(\contVal)] \cdot \phi_{\discVal}.
\end{equation}

From~\cref{eq:appndx-policy-iter}, the two terms of interest for the policy iteration are the expected cost of the policy $\hat{J}_{\discVal}^k(c \mid \pi)$
and the probability of failure to reach the target $[1 - \rho_{\pi}(\contVal)]$. The above analysis explicitly shows
how $\phi_{\discVal}$ is a scaling factor that balances these two terms. By setting it to the quantity in~\cref{eq:terminalcost},
we are effectively prioritizing policies that reach the target in time over ones that do so with lower expected cost within the region.
Ultimately, this choice of a penalty term is still a heuristic.

\section{Global Layer Heuristic}
\label{sec:appendix-heuristic}

The heuristic function $\mathrm{heur}$ in~\cref{alg:HSP-global} is used by the global layer planning to guide the search 
over good mode sequences and (hopefully) make it more efficient than searching over all possible mode
sequences, which may be unacceptably expensive. Heuristic functions in search~\cite{pearl1985heuristics} 
are usually goal-directed (usually an easy-to-compute estimate of the cost to reach the goal from the
current state).
They also usually operate on points, not spaces.
Therefore, a heuristic function of the form $\mathrm{heur}(s_t,\calS_G)$ where the second argument is a goal space will
not necessarily be well-defined. Of course, domain-specific heuristics may be able to work with a goal space (for instance by sampling
a goal state), but we cannot
make a general comment on that.

Therefore, in~\cref{alg:HSP-global}, the heuristic operates only on the set of modes, i.e.
$\mathrm{heur} : \discSet \times \discSet \rightarrow \mathbb{R}^{+}$, and for a given state $s_t = (\discVal_t, \contVal_t)$,
the heuristic value is $\mathrm{heur}(\discVal_t, \discVal_G)$ where $\discVal_G$ is the \emph{goal mode}, i.e.
the mode of the goal space.
There has been a long line of work on domain-agnostic heuristics for classical planning~\cite{pearl1985heuristics,hoffmann2001ff,helmert2009landmarks},
many of which can be utilized here while searching over the logical modes.
As we mentioned in~\cref{sec:appendix-opt-global}, global optimality guarantees cannot be made for DMSSPs, so the heuristic
functions need not be admissible, i.e. under-estimates of the cost to reach the goal. In any case, due to the uncertainty in the
outcome of the underlying control actions, it can in general be challenging to devise useful admissible heuristics in stochastic settings.
A potentially useful non-admissible heuristic could be based on a worst-case traversal cost within modes.

\section{Horizon Limit Selection}
\label{sec:appendix-horizon}

In the main text, for simplicity, we assumed that the horizon limit used is the same as the
context horizon limit $K$, which is an appropriate choice if that value is known
for a particular domain. In this section we discuss some general issues about the parameter
and one possible choice that does not require knowledge of the context horizon limit.
For the subsequent discussion, we denote the local layer's horizon limit parameter as $K_{\discVal}$ to
distinguish it from the context horizon $K$.
The choice of horizon limit parameter for the local layer influences the lookahead of the local MDP policies and the
amount of memory storage required for the value function (increases linearly with $K_{\discVal}$).

A very small limit will make the local layer more sensitive
to the probability distribution over context horizon $P_{1:K}$ for the next mode switch.
In our conveyor belt example, suppose the next planned mode switch is to pick up a box
at a point on its future expected
trajectory. If $K_{\discVal} \ll K $, then for most future points on the box's trajectory, we will have
$\sum_{k > K_{\discVal}}^{K} P(k) \sim 1$, i.e. the bulk of the probability mass is on
future horizon values greater than the local layer horizon limit. The local layer cannot then choose
a useful control action from its cost-to-go function $J_{\discVal}^{0:K_{\discVal}}$ in~\cref{eq:region-policy}
(the $K$ from~\cref{eq:region-policy} is $K_{\discVal}$ in this case). It has to wait until
$\sum_{k \leq K_{\discVal}}^{K} P(k)$ is sufficiently greater than $0$, but that restricts the total reaction time
of the local layer. Thus, it can only choose transition points up to a few timesteps ahead, which 
reduces the robustness to downstream effects, i.e. missing time-constrained mode switches.

A very large horizon value will accommodate the context horizon but will make the value function $\costtogo$ expensive to
compute and store. Subsequently, we propose a domain-agnostic strategy that does not require knowledge of the context horizon
limit. We make the same assumptions on the local intra-modal MDP as in~\cref{sec:appendix-opt-local}.
Also denote $\lVert \contVal \rVert$, i.e. the norm of the relative state as the distance to the origin; the target is
reached when $\lVert  \contVal  \rVert \leq \epsilon$ for some $\epsilon$. The intuitive idea is to set $K_{\discVal}$ 
at least high enough to allow any relative state to reach the origin
with some set of control actions. 

Let  $\bar{\calA}_{\contSpace} = \calA_{\contSpace} \setminus \text{no-op}$ be the set of all control
actions excluding the no-op action.
We define the \emph{worst-case progress} of an action $a \in \bar{\calA}_{\contSpace}$ as
\begin{equation}
\label{eq:appendix-prog}
\rho(\contVal, a) = \max_{\contVal' \in \contSpace} \frac{\lVert \contVal' \rVert}{\lVert \contVal \rVert} \ \text{where} \ 
T_{\discVal}(\contVal, a, \contVal') > 0
\end{equation}
where $\rho(\contVal, a) < 1$ indicates that the action makes progress towards the origin, 
as the distance to the origin after the action has reduced. for all possible next states. Another \emph{assumption} we make is
that progress towards the origin can be made from every state, i.e.
\begin{equation}
\label{eq:appendix-assumption}
\forall \contVal \in \contSpace, \ \exists \ a \in \bar{\calA}_{\contSpace} \ \text{such that} \ \rho(\contVal, a) < 1
\end{equation}
and the set of all such progress actions for a given relative state is denoted $\bar{\calA}^{\rho}(\contVal)$.
Furthermore, define the \emph{most progressive action} for a relative state as the one with maximum worst-case progress, i.e.
\begin{equation}
\label{eq:appendix-most-progress}
a_{\rho}^{*}(\contVal) = \underset{a \in \bar{\calA}^{\rho}(\contVal)}{\mathrm{argmax}} \ \rho(\contVal, a).
\end{equation}
Finally, let the \emph{minimum progress} from any state in one step be denoted as
\begin{equation}
\label{eq:appendix-min-prog}
\delta = \min_{\contVal \in \contSpace} \rho\left(\contVal, a_{\rho}^{*}(\contVal)\right).
\end{equation}
By the assumption in~\cref{eq:appendix-assumption} and by~\cref{eq:appendix-most-progress}, we know that $\delta < 1$.
From any relative state, there is always an action that can reduce the distance to the origin by a fraction of \emph{at least} $\delta$.
Therefore, a suitable choice for $K_{\discVal}$ satisfies
\begin{equation}
\label{eq:appendix-hor-lim}
\delta^{K_{\discVal}} < \epsilon \implies \framebox{$K_{\discVal} = \lfloor \log_{\delta} \epsilon \rfloor$}
\end{equation}
where $\epsilon$ is the domain-dependent threshold parameter.

The above analysis of~\crefrange{eq:appendix-prog}{eq:appendix-hor-lim} does assume 
that the various involved quantities are computable for a continuous space $\contSpace$.
For simple dynamical systems it may be possible to do so analytically without sampling states from the space,
otherwise, a sampling scheme can be used during preprocessing to generate an exhaustive set of samples
and then we can directly apply the equations to compute a horizon limit
parameter which is sufficient for most of the space.

\section{Implementation Details}
\label{sec:appendix-impl}

\textbf{N.B.} Due to legacy naming reasons, the attached code uses the acronyms `CMSSP' and `HHPC' instead of `DMSSP' and `HSP' respectively.

As we mentioned earlier, our implementation is in the Julia programming language. We also rely heavily on the POMDPs.jl~\cite{egorov2017pomdps}
framework for modeling and solving Markov Decision Processes.
Our implementation broadly consists of: (i) a \emph{domain-agnostic} component which defines the DMSSP problem model
and an interface for defining the various components,
and the general HSP solution framework;
(ii) a \emph{domain-specific} component which instantiates the various DMSSP components (discrete modes 
and continuous state space, transition, reward, context) and the other functions required by the HSP solution
framework (\textsc{NextValidModes}, \textsc{SampleTransition}, and so on). We briefly describe the domain-agnostic component here
and the domain-specific component (for multimodal routing) in~\cref{sec:appendix-exp-impl}.
We provide more elaborate technical details in the README of the attached code.

The DMSSP problem formulation is specified in the \texttt{src/models/} folder. It is primarily an interface that is parameterized
by the domain-specific datatypes for the state and action spaces and the context set, that the domain designer has
to provide. The HSP solution framework is implemented in the \texttt{src/hhpc/} folder. As for the algorithms in the paper, it is
divided into a \texttt{global\_layer}, a \texttt{local\_layer}, and the full \texttt{hhpc\_framework}.

The \texttt{global\_layer} implements the \textsc{GlobalPlan} procedure of~\cref{alg:HSP-global} using a modified A* search algorithm implementation~\cite{hart1968formal}. 
The modification is for \emph{implicit} graphs, i.e.
where the edges are not specified before the search is called, but rather generated on-the-fly by a successor function
when a node is expanded. An implicit search allows greater flexibility, \emph{especially in domains with many discrete modes}; 
an efficient \textsc{NextValidModes} subroutine can be used to only generate the modes actually reachable from the current mode,
rather than explicitly enumerating all of them apriori. The \texttt{local\_layer} implements the \textsc{LocalPreprocessing} procedure of~\cref{alg:HSP-global}, i.e.
the terminal cost computation and finite-horizon approximate value iteration. We use multilinear
grid interpolation~\cite{busoniu2010reinforcement} of the value function over the continuous state space, however, an alternate 
implementation with a different approximator could also be done here.

The \texttt{hhpc\_framework} defines the top-level behavior of~\cref{alg:HSP-full}. It assumes access to a discrete-time domain-specific
DMSSP simulator. At every time-step, it observes the current state and context set, makes a decision based on the HSP framework, and outputs 
an action to the simulator. It also implements the hierarchical interleaving of planning at the global layer
and execution at the local layer.

\section{Further Experimental Details}
\label{sec:appendix-exp}

The experimental domain used is the recently introduced dynamic real-time multimodal routing (DREAMR)
problem~\cite{choudhury2019dynamic}.
An agent (for example a drone) has to be controlled from a start to a goal location.
There is a network of transit vehicle routes that the agent has access to, and it may use
transit vehicles as temporary modes of transportation along segments of the routes,
in addition to moving on its own to the destination. The objective is to reach
the goal location while incurring as low a trajectory cost as possible,
where the agent is penalized for energy expended due to distance traversed and hovering
in place, and total elapsed time.

In this section, we provide a more elaborate description of the experiments we ran to evaluate HSP 
against the two baseline methods of two-level Receding Horizon Control (RCH) and Upper Confidence Trees (UCT)
with enhancements from PROST and Double Progressive Widening.
Several implementation aspects pertaining to the problem domain of Dynamic Real-time Multimodal Routing (DREAMR) were 
obtained from the open-source Julia repository DreamrHHP.jl (\href{https://github.com/sisl/DreamrHHP}{here}) of the original paper.
All the experiments were run on Linux with \SI{16}{\gibi\byte} RAM and a $6$-core \SI{3.7}{\giga\hertz} CPU.

\subsection{Domain-Specific Implementation}
\label{sec:appendix-exp-impl}

On the DMSSP problem formulation side, the state, action, transition, and reward functions are all obtained directly from the DREAMR paper. The context is
the current position and estimated remaining route (as a sequence of time-stamped waypoints) for all currently active
transit vehicles. The information is summarized below:

\begin{align*}
    &\discVal_t \in \{\textsc{Move}, \textsc{Ride}\} \ ; \ \contVal_t = \left(x_t,y_t,\dot{x}_t,\dot{y}_t\right) \ ; \ s_t = \left(\discVal_t, \contVal_t \right) &\text{State Space} \\
    &a_{\discSet} \in \{\textsc{Board}, \textsc{Alight}\} \ ; \ a_{\contSpace} = \left(\ddot{x},\ddot{y}\right) &\text{Action Space} \\
    &\context_t = \left(\bfp_{t}^{1}, \mathbf{w}_{t}^{1}, \bfp_{t}^{2}, \mathbf{w}_{t}^{2}, \ldots \bfp_{t}^{n_t}, \mathbf{w}_{t}^{n_t}\right) &\text{Context set} \\
    &\bfp_{t}^{i} = \text{Current position of vehicle $i$} \\
    &\mathbf{w}_t^{i} = ((\bfp_{1,t}^{i},\tau_{1,t}^{i}),(\bfp_{2,t}^{i},\tau_{2,t}^{i}), \ldots) &\text{ETA stamped future waypoints} \\
    &\contVal_{t+1} = f(\contVal_t,a_{\contSpace,t} + \epsilon), \ \epsilon \sim \mathcal{N} \left(\mathbf{0}, \mathrm{diag}(\sigma_{\ddot{x}}, \sigma_{\ddot{y}})\right) &\text{Control Dynamics}\\
    &T(\textsc{Move},\contVal_t,a_{\contSpace,t}) \sim f(\contVal_t,a_{\contSpace,t} + \epsilon) &\text{\textsc{Move} mode dynamics}\\
    &T(\textsc{Ride},\contVal_t,a_{\contSpace,t}) \sim \context_{t+1} &\text{Determined by ride vehicle position}\\
    &T(\context_t, (\textsc{Move},\contVal_t),\textsc{Board}) = (\textsc{Ride}, \contVal_{t+1}) &\text{If $\contVal_t$ close to car and speed close to 0}\\
    &T(\context_t, (\textsc{Ride},\contVal_t), \textsc{Alight}) = (\textsc{Move}, \contVal_{t+1}) &\text{Can alight anytime from a vehicle}\\
    &R(s_{t},a,s_{t+1}) = -(\underbrace{\lambda_d \lVert \contVal_{t+1} - \contVal_{t} \rVert_2 + \lambda_h \mathds{1}_{h}}_{\mathrm{energy}} +  \underbrace{1}_{\mathrm{time}}) &\text{Cost function}
\end{align*}
where $\mathds{1}_h = \mathds{1}[\lVert (\dot{x_t}, \dot{y_t}) \rVert < \epsilon]$ indicates hovering in place, $\lambda_d$ and $\lambda_h$ are scaling parameters
for how important the distance traversed and hovering are with respect to each other and with respect to each unit of elapsed time.

On the HSP solution side, there are only a few details worth mentioning. For simplicity, the transition 
between modes is constrained to only happen at the transit vehicle route waypoints. For \textsc{Move} to
\textsc{Ride}, the agent can \textsc{Board} at potentially any of the future route waypoints of all the currently active cars,
while for the converse, the agent can \textsc{Alight} at any of the future route waypoints of the transit vehicle it is currently on.
We make this simplification because, at least for this work, we are not interested in tuning the sampling parameter $N$
and evaluating its effect on the performance of HSP. In practice, we can certainly increase resolution by sampling transition points
in between the pre-decided route waypoints. In any case, \emph{because the ETA at the route waypoints is perturbed with high probability
at each timestep, we simulate dynamically changing contextual information.}

The agent's $(x,y)$ position is assumed to be bounded on a $1 \times 1$ unit grid (that can be arbitrarily scaled to represent a real-world grid).
Since the agent's control is holonomic, and it can move in any direction, we use the \emph{relative position} to encode
the local layer MDP state; accordingly, the XY bounds are $[-1,1] \times [-1,1]$.
As we subsequently mention in~\cref{sec:appendix-exp-probs}, the problem scenarios simulate routes on a grid representing \SI{10}{\kilo\metre} $\times$ \SI{10}{\kilo\metre},
therefore velocity and acceleration limits for the agent (drone) are scaled accordingly to reflect real-world limits. 
All parameter values are based on the original set of values used for the DREAMR experiments, and detailed
in \texttt{DreamrHHP/data/paramsets} of the \href{https://github.com/sisl/DreamrHHP}{DreamrHHP.jl} repository.
For example, in \href{https://github.com/sisl/DreamrHHP/blob/master/data/paramsets/scale-1.toml}{scale-1.toml}, the
\texttt{XYDOT\_LIM} parameter which sets the speed threshold in each direction is $0.002$ representing
\SI{20}{\metre\per\second}, which is the maximum speed of the \href{https://www.dji.com/phantom-4/info}{DJI Phantom 4}.

\subsection{Problem Scenarios}
\label{sec:appendix-exp-probs}

As we mentioned in~\cref{sec:experiments-results}, we use the exact same problem scenarios from the original
DREAMR paper (\cite{choudhury2019dynamic}, sec. V-B), and the following description
is largely derived from the reference work.
Since we care about the higher-level decision making framework, we abstract away physical issues like obstacles, collisions,
and so on.
The $1 \times 1$ unit grid represents an area of \SI{10}{\kilo\metre} $\times$ \SI{10}{\kilo\metre}
(approximately the size of north San Francisco). Each episode lasts for $30$ minutes, with $360$ timesteps or epochs of \SI{5}{\second} each. An episode starts with between $50$ to $500$ cars, 
with more added randomly at later epochs (up to twice the initial number). Therefore the total number of cars
over the episode is $100$ to $1000$.  

A new car route is generated by first choosing two endpoints more than $\SI{2}{\kilo\metre}$ ($0.2$ units) apart. We choose a number of route waypoints from a uniform distribution of $5$ to $15$, denoted $\mathrm{U}(5,15)$, and a route duration from $\mathrm{U}(100,900)$s. Each route is either a straight line or an L-shaped curve. The waypoints are placed along the route with an initial ETA based on an average car speed of up to \SI{50}{\metre\per\second}. 
At each epoch, the car position is propagated along the route and the ETA at each remaining waypoint is perturbed with $p = 0.75$ within $\pm$\SI{5}{\second} (highly dynamic). The route geometry is simple for implementation, but we represent routes as a sequence of waypoints. This is \emph{agnostic to route geometry or topology} and can accommodate structured and unstructured scenarios. In all problems, the agent begins at the centre of the grid, and the goal is near a corner.

\begin{table}[t]
\centering
\begin{tabular}{@{} rrrrr @{}}
    \toprule
    Variant & depth & exploration & n\_iterations & init\_N \\
    \midrule
    UCT1        & \num{100}   & $50.0$ & $500$   & $1$\\
    UCT2        & \num{200}    & $1.0$ & $500$   & $1$\\
    UCT3        & \num{100}   & $100.0$ & $500$    & $50$\\
    \bottomrule
\end{tabular}
\caption{The relevant parameter values for the three variants of the UCT baseline.}
\label{table:appendix-uct}
\end{table}

\subsection{Further Evaluation Details}
\label{sec:appendix-exp-eval}

We provide some additional details on the baselines, compare the various
methods on two additional sets of problems of the DREAMR domain, and provide
average computation times.

\subsubsection{Baselines}

We use the same two-level Receding Horizon Control baseline that is domain-specific
to the DREAMR problem from the original paper. We used the 
open-source implementation from their accompanying repository (\href{https://github.com/sisl/DreamrHHP/blob/master/scripts/mpc_solver.jl}{link}).
For UCT,~\cref{table:appendix-uct} lists the
relevant parameters for the three variants of UCT which are used as a baseline in~\cref{sec:experiments-results}. The parameters are based on the open-source implementation (\href{https://github.com/JuliaPOMDP/MCTS.jl/blob/master/src/dpw_types.jl}{link}) of
Double Progressive Widening that we use. 
We had a wider full range of parameters (14 different sets) and we chose the three most varying and representative ones.
For all UCT variants, we used the default values for the double progressive widening branching parameters
$(k_a = 10, \alpha_a = 0.5, k_s = 10, \alpha_s = 0.5)$

\begin{table}
\centering
\begin{tabular}{@{} lrr @{}}
    \toprule
    Episode Set & Num. Routes & Perturbation Prob. \\
    \midrule
    Set 1 (in main) & $\SIrange[range-phrase = -]{100}{1000}{}$ & $0.75$ \\
    Set 2   & $\SIrange[range-phrase = -]{500}{2500}{}$ & $0.75$ \\
    Set 3   & $\SIrange[range-phrase = -]{500}{2500}{}$ & $0.35$\\
    \bottomrule
\end{tabular}
\caption{The three sets of problems (with $1000$ episodes each) we used for evaluating the performance of our HSP approach against the two baselines.
The two parameters varied are the number of active car routes and the probability of a perturbation (speedup/delay) to
the ETA of a future route waypoint at each timestep. As an additional note, each route has an average of $10$ waypoints; multiplying the number of routes
by $10$ will give an idea of the number of route vertices being considered.}
\label{table:appendix-problemsets}
\end{table}

\begin{figure}
    \centering
    \begin{subfigure}{0.49\textwidth}
        \centering
        \includegraphics[width=\textwidth]{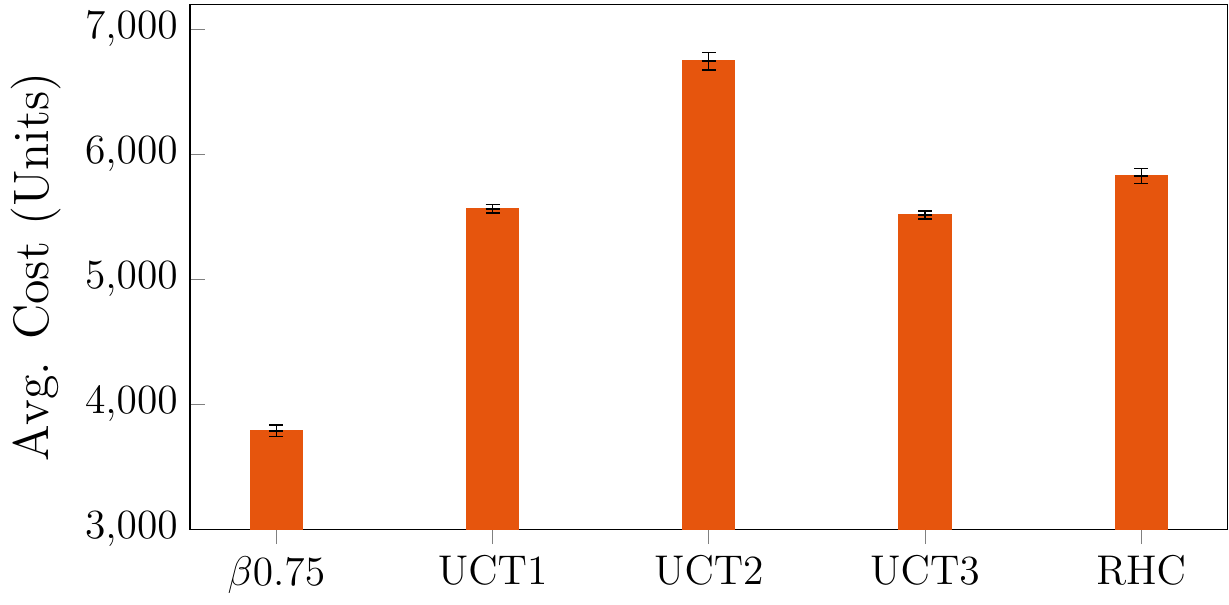}
        \caption{Set 1 (From main)}
        \label{fig:avg-set1}
    \end{subfigure}
    \begin{subfigure}{0.49\textwidth}
        \centering
        \includegraphics[width=\textwidth]{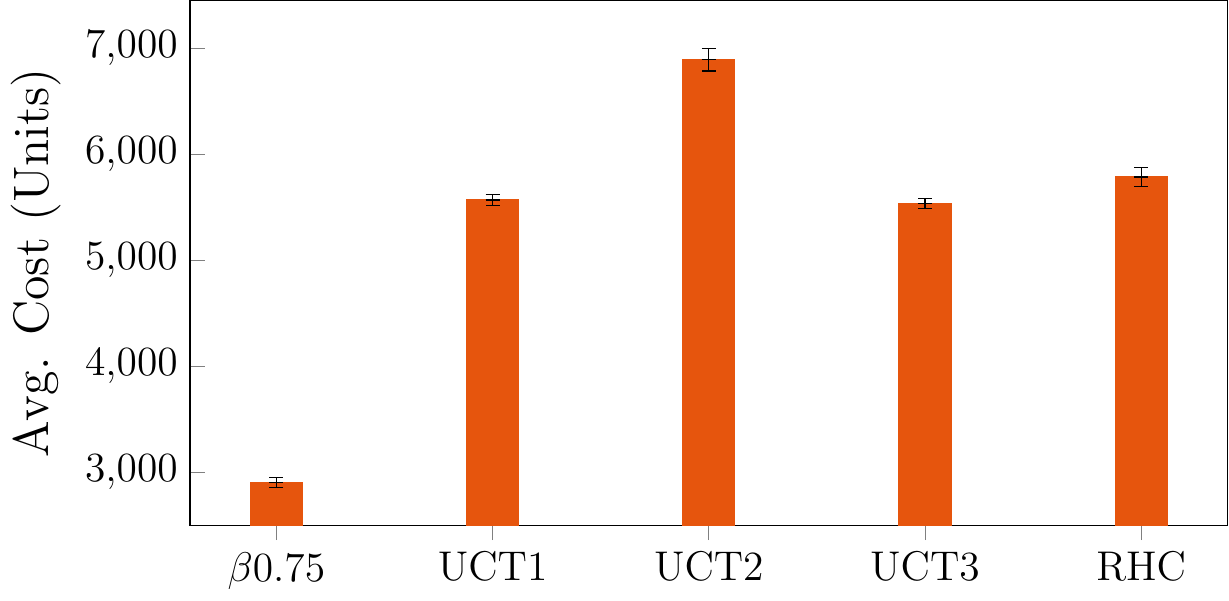}
        \caption{Set 2}
        \label{fig:avg-set2}
    \end{subfigure}

    \begin{subfigure}{0.49\textwidth}
        \centering
        \includegraphics[width=\textwidth]{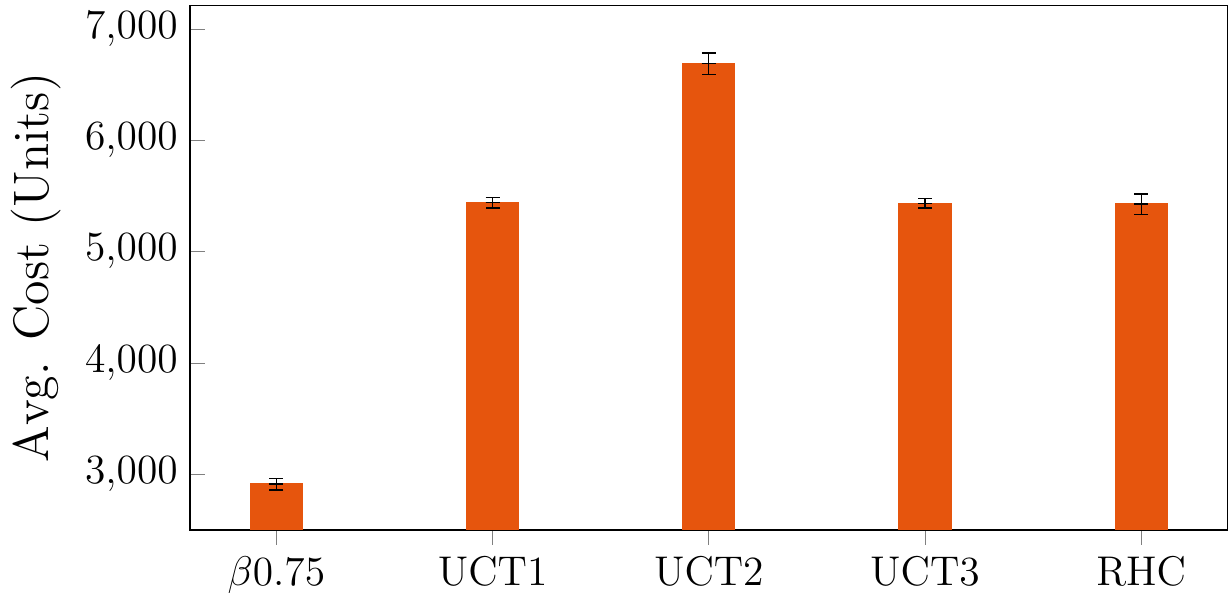}
        \caption{Set 3}
        \label{fig:avg-set3}
    \end{subfigure}
    \caption{Average cumulative trajectory cost to reach the goal, over $1000$ episodes each for the three sets of problems
    described in~\cref{table:appendix-problemsets}. Error bars are for standard error.
    Set 1 (\cref{fig:avg-set1}) is reproduced from~\cref{fig:cost-time} of the main paper here for a comparison. For HSP, we only show the $\beta = 0.75$
    results. The relative improvement of HSP over the baselines
    is higher in Sets 2 and 3 (b and c) as the number of transit routes is much higher than in Set 1 and using them judiciously can lead to more energy savings. Also, there is a slight decrease
    in cost incurred for most methods between Sets 2 and 3 as the waypoint ETA perturbation probability is less. However, HSP's performance is more invariant to the perturbation parameter than
    the baselines, implying higher robustness to the variation in the context set.}
    \label{fig:avg-sets}
\end{figure}

\subsubsection{Two Additional Episode Sets}
\label{sec:appendix-exps-addtnl}

The two important parameters that define a DREAMR problem scenario are the number of active car routes over the episode (a measure 
of the size of the context set and the number of valid mode sequences to the goal)
and the probability of perturbing the ETA at remaining route waypoints within $\pm$\SI{5}{\second} (a measure of how dynamic the context set is). For the results in~\cref{sec:experiments-results},
of the main paper, we generated $1000$ episodes where the number of cars is $100$ to $1000$ and the waypoint perturbation probability is $p = 0.75$.
This replicates the set of scenarios from the original DREAMR paper. Additionally, we generate two other sets of $1000$ episodes; the parameter values for all three sets are depicted in~\cref{table:appendix-problemsets}.

For all three problem sets, the average trajectory cost for each method to reach the goal is depicted in~\cref{fig:avg-sets} and the average number of mode switches in~\cref{table:modeswitch-sets}. The procedure for obtaining the statistics is the same as in~\cref{sec:experiments-results} of the main paper. For HSP we only plot the values for $\beta = 0.75$
as the values for $0.55$ and $0.95$ are nearly identical to it, as was the case
for Set 1. The plots demonstrate that HSP consistently outperforms the baselines
.They also show how the relative performance gap between HSP and the baslines increases with more transit vehicle routes (from Set 1 to Sets 2-3), i.e. more valid mode sequences and a greater benefit in energy saved for choosing good connections and making them in time.

Furthermore, between Sets 2 and 3 (which have the same transit vehicle routes for each episode but Set 2 has a higher ETA perturbation probability), there is a slight decrease in cost incurred for most algorithms
in Set 3 compared to Set 2 (as well as a slight increase in the number of mode switches). This is not unexpected as the context is changing more dynamically in Set 2, potentially increasing the number of missed connections or time spent hovering due to a delay.
For HSP, the relative performance change between Sets 2 and 3 is minimal, far lower than the relative change for the other algorithms. Specifically, between Sets 2 and 3, the relative decrease in cost for the UCT variants is roughly $2.5\%$ and for RHC it is roughly $6.6\%$ (deterministic replanning
is the most sensitive to perturbation),
but for HSP it is only $0.2\%$. \emph{Therefore, HSP is the most robust
to the variation in the context set}, manifested here as the waypoint time-stamp perturbation.

\begin{table}[t]
\centering
\begin{tabular}{@{} rrrr @{}}
    \toprule
    Algorithm & Set 1 & Set 2 & Set 3 \\
    \midrule
    $\beta0.75$ (HSP)   & $3.422$   & $4.305$   & $4.399$ \\
    UCT1                & $0.024$   & $0.082$   & $0.096$ \\
    UCT2                & $0.018$   & $0.037$   & $0.032$ \\
    UCT3                & $0.015$   & $0.261$   & $0.289$\\
    RHC                 & $3.508$   & $4.664$   & $4.782$\\
    \bottomrule
\end{tabular}
\caption{The average number of mode switches for the various algorithms on the three
problem sets, where Set 1 is reproduced from the main body for completeness . There is a sharp relative increase from Set 1 to Set 2 as the
number of transit routes increases significantly. For most algorithms, there is also
a slight increase from Set 2 to Set 3 as the waypoint perturbation probability is lower,
i.e. the context changes less dynamically, allowing slightly more connections to be made.}
\label{table:modeswitch-sets}
\end{table}

% Set 1
% hhpc75_tuple = (4403.671035683799, 47.760250125595874, 241.06021251475798, 1.933630855197148, 3.64344746162928)
% dpw1_tuple = (6588.216095481236, 27.459565706412203, 113.97520661157024, 0.7157530758418569, 0.0590318772136954)
% dpw2_tuple = (8342.054900458565, 75.73184957534278, 157.43565525383707, 1.668763802066627, 0.021251475796930343)
% dpw3_tuple = (6499.157968533448, 24.55222906107958, 106.9456906729634, 0.6532553553484189, 0.06375442739079103)
% mpc = 6217.915, 55.0775, 3.634

% Set 3
% hhpc75_tuple = (2905.8259409236757, 47.69640717647024, 231.7995444191344, 2.73666118544768, 4.305239179954442)
% dpw1_tuple = (5567.9915271102045, 51.131057683478744, 97.53424657534246, 1.1218627097969134, 0.0821917808219178)
% dpw2_tuple = (6893.279889033335, 106.71079410310509, 129.78995433789953, 2.2613697325046638, 0.0365296803652968)
% dpw3_tuple = (5535.657697299563, 45.49239888665918, 92.09132420091325, 0.9154808090913134, 0.2602739726027397)
% rhc = 4.664

% Set 4

% hhpc75_tuple = (2911.214950743943, 50.69510779555419, 236.8348623853211, 2.6824896776057834, 4.39908256880734)
% dpw1_tuple = (5438.5613664732455, 47.123832605893455, 96.54357798165138, 1.0768385248379044, 0.0963302752293578)
% dpw2_tuple = (6687.695391656403, 95.85459219800225, 125.92201834862385, 2.013750250086638, 0.03211009174311927)
% dpw3_tuple = (5434.217084968327, 43.56101242877347, 91.62155963302752, 1.0028971986158446, 0.2889908256880734)
% mpc 4.782

\begin{table}
\centering
\begin{tabular}{@{} rr @{}}
    \toprule
    Algorithm & Avg. Time (\SI{}{\milli\second}) \\
    \midrule
    HSP Global Layer & $\SIrange[range-phrase = -]{100}{200}{}$\\
    HSP Local Layer & $\SIrange[range-phrase = -]{10}{40}{}$\\
    UCT1 & $\SIrange[range-phrase = -]{80}{90}{}$\\
    UCT2 & $\SIrange[range-phrase = -]{85}{95}{}$\\
    UCT3 & $\SIrange[range-phrase = -]{90}{125}{}$\\
    RHC Global Layer & $\SIrange[range-phrase = -]{100}{150}{}$\\
    RHC Local Layer & $\SIrange[range-phrase = -]{8}{30}{}$\\
    \bottomrule
\end{tabular}
\caption{ The problem set used for evaluating the computation is the first set of episodes, where the average
number of car routes is $100$ to $1000$ for an episode. The decision frequency of HSP (that of its local layer except after asynchronous interrupts) is  comparable
to that for RHC and slightly faster than UCT. The global planning layer of HSP is slightly slower than that RHC's global layer (the cost-to-go lookup is a bit more expensive than the nominal edge weight). }
\label{table:appendix-times}
\end{table}

\subsubsection{Computation Times}

For the results in~\cref{sec:experiments-results}, in the main body, we focused on solution quality as our primary metric.
However, we were also motivated to mitigate the increase in complexity due to considering uncertainty in our
hybrid planning framework.~\Cref{table:appendix-times} compares the computation time for our framework
against the baseline methods. 
For HSP and RHC, we show planning times for both the global layer and the local 
layer. In practice, as we mentioned in~\cref{sec:hsp-interleaving}, after the first plan, the global layer
could plan periodically in the background while the local layer executes the current plan. Therefore, in general,
the decision frequency of the HSP framework is that of the local layer. Only when there is an
asynchronous or event-driven interrupt, would the HSP framework be bottlenecked by the global layer.
For UCT, we show the computation time required to select the next action (there are no layers).

The computation times were obtained by randomly choosing a subset of the $1000$ episodes \emph{of Set 1 (where the number 
of car routes is between $100$ and $1000$)} and computing the average
elapsed time for the various methods. The computation time depends on the exact size of the context set for the episode, so we provide
an approximate range of values. Also, for the global layer, we only use the computation time for the first $25\%$ searches 
(subsequent global searches become trivially fast as the agent gets closer to the goal).
As~\cref{table:appendix-times} demonstrates, compared to RHC, which repeatedly solves a deterministic problem, we are only slightly less efficient computationally, at both global
and local layers. Compared to UCT, the local layer of HSP is at least two times faster, which we do expect
as the local layer is looking up a policy computed offline while UCT is doing online planning.
Even the global planning of HSP, which has to  search over mode sequences and transition points, is only up to
two times slower than UCT.

\end{document}